\documentclass[preprint,12pt,authoryear]{article}

\usepackage{blindtext}
\usepackage{amssymb}
\usepackage{amsmath}
\usepackage{amscd}
\usepackage{amsfonts}
\usepackage{amssymb}
\usepackage{amsthm}
\usepackage{array}
\usepackage{graphicx}
\usepackage{wrapfig}
\usepackage{url}
\usepackage{textcomp}
\usepackage{textfit}
\usepackage{color}
\usepackage[table]{xcolor}
\usepackage{setspace}
\usepackage{booktabs}
\usepackage{algpseudocode}
\usepackage{textfit}
\usepackage{bm}		
\usepackage{bbm}					
\usepackage{multirow}
\usepackage{fancyhdr}
\usepackage{booktabs}
\usepackage{placeins}
\usepackage{bm}
\usepackage{numprint}
\usepackage{arydshln}
\usepackage{pifont}
\usepackage{bibentry}
\usepackage{lscape}
\usepackage{tikz}
\usepackage{multicol}
\usetikzlibrary{decorations}
\usepackage{float}

 \usepackage{amsthm}
 
 \newtheorem{lemma}{Lemma}

\begin{document}




\title{Context-specific kernel-based hidden Markov model for time series analysis}


%
%
%
%
%
%
%

\author{ Carlos Puerto-Santana$^{1,2}$,  
	Concha Bielza$^2$,             
	Pedro Larrañaga$^2$ and    \\       
	Gustav~Eje~Henter$^3$           
 }

\date{%
	\small{
	$^1$Aingura IIoT, San Sebastian, Spain\\%
	$^2$Computational~Intelligence~Group, Technical~University~of~Madrid, Madrid, Spain\\%
    $^3$Division~of~Speech,~Music~and~Hearing, KTH~Royal~Institute~of~Technology, Stockholm, Sweden}\\[2ex]%
	\today }

\maketitle
\begin{abstract}
Traditional hidden Markov models have been a useful tool to understand and model stochastic dynamic data; in the case of non-Gaussian data, models such as mixture of Gaussian hidden Markov models can be used.
However, these suffer from the computation of precision matrices and have a lot of unnecessary parameters.
As a consequence,  such models often perform better when it is  assumed that all variables are independent, a hypothesis that may be unrealistic.
Hidden Markov models based on kernel density estimation are also capable of modeling non-Gaussian data,  but they assume independence between variables.
In this article, we introduce a new hidden Markov model based on kernel density estimation, which is capable of capturing kernel dependencies using context-specific Bayesian networks.
The proposed model is described, together with a learning algorithm based on the expectation-maximization algorithm.
Additionally, the model is compared to related HMMs on synthetic and real data.
From the results, the benefits in likelihood and classification accuracy from the proposed model are quantified and analyzed.

\end{abstract}

\section{Introduction}

Nowadays, the analysis and understanding of stochastic dynamic data have obtained more attention in areas such as industry 4.0, weather forecasting, facial and speech recognition, biology, economy and so on.
In real data, hypotheses such as homoscedasticity (constant variance) or independence between input variables  do not hold and traditional models for continuous data, such as auto-regressive (AR) or moving average (MA) linear regression (or combinations of these), are often unsuitable.
Dynamic probabilistic graphical models such as hidden Markov models (HMMs) \cite{1990:rabiner}  can alleviate such issues and  be used to provide further data insights, compute likelihoods, perform data segmentation or classification.
However, traditional HMMs,  are usually based on the Gaussian distribution which limits their modeling capabilities to Gaussian data.
To overcome this issue, HMMs with mixture of Gaussians (MoGs) have been proposed (as for instance by \cite{1985:juang.rabiner}); nonetheless, due to numerical issues regarding underflow/overflow when computing inverses of covariance matrices, such models have to assume diagonal covariance matrices, i.e., independence among the model features.
More recently, asymmetric HMMs (As-HMMs) (see \cite{2017:bueno.hommersom.ea,2022:puerto-santana.larrnaga.ea}) have been proposed to model variable dependencies in a more stable computational  manner and using fewer parameters than in the case of MoG-HMMs, via the use of context-specific Bayesian networks (BNs), or BNs which change their structure depending on the value of a certain variable, in the emission probabilities, see \cite{1996:boutilier.friedman.ea}.
However, such models usually assume Gaussian data, which limit their applicability. 

In this article, we propose a new kind of As-HMMs where non-Gaussian dynamic data is modeled using kernel density estimation (KDE) (see \cite{1956:rosenblatt}).
By adding context-specific dependencies between variables to the model, As-HMMs are made more expressive.
We call the models \textit{kernel density estimation in asymmetric hidden Markov models} or KDE-AsHMMs.

We validate the new model by applying it to synthetic data and to real data from ambient sound and from CNC drilling machines, comparing model capabilities in terms of classification accuracy and likelihood to previous HMM-based models
Additionally, theoretical bounds on computational time complexity on the learning and inference (log-likelihood computation) algorithms are provided, for a better characterization of the time cost when these new models are applied.

In short, the present paper provides the following contributions to the state of the art related to HMMs:
\begin{itemize}
	\item[1.] A new kind of asymmetric HMMs are introduced to model non-Gaussian data based on KDEs and context-specific BNs.
	\item[2.] All the parameters are interpretable and can provide further data insights.
	\item[3.] A learning algorithm, based on the expectation-maximization (EM) and structural EM algorithms, is provided to learn KDE-AsHMMs.
	\item[4.] Complexity bounds on computation time are provided for further model behavior  understanding.
\end{itemize}

The paper is organized as follows. 
Section~\ref{sec:asota} surveys the state of the art related to BNs in HMMs and KDEs. 
Section~\ref{sec:background} gives a brief summary of the main ideas used in the article.
Section~\ref{sec:proposed} presents KDE-AsHMM, their learning algorithm and theoretical time-complexity upper bounds.
Section~\ref{sec:experiments} describes the validation and relevant comparisons with synthetic and real data. 
Finally, Section~\ref{sec:conclusions} rounds-off the article with the conclusions and considerations based on the model definition and validation findings.
The model code and the experimental set-up scripts can be found online at: \url{https://github.com/Puerto-Santana/PyAsHMM}.

\section{Related work}\label{sec:asota}

The contribution of this paper consists of introducing a new family of HMMs,  where the emission probabilities are expressed with KDEs and context-specific BNs.
Therefore, we first review the relevant bibliography related to KDEs with BNs and HMMs.

\subsection{Bayesian networks}

Regarding BNs, \cite{1995:hofmann.tresp}  proposed a heuristic to learn BNs with kernel conditional estimation (KCE).
The networks were learned using leaving-one-out scores penalized by the number of arcs in the model. 
In this model, each feature had its own bandwidth parameter to be estimated. 
\cite{2009:perez.larranaga.ea} introduced kernel-based BNs for classification, where the bandwidths were computed using rule of thumb formulas, see \cite{1986:silverman}.
They also incorporated different methodologies for supervised graph structure into their model.
It was found through statistical testing that the tree augmented algorithm obtained the minimum prediction error for supervised datasets. 
Finally, \cite{2022:atienza.bielza.ea} proposed a semi-parametric BN where the dependencies between variables were defined in a linear Gaussian Bayesian networks fashion or KCE. 
The authors proposed a graph search algorithm based on the tabu meta-heuristic, see \cite{1986:glover}; in the graph search, each node could change its dependency model (linear Gaussian or kernel-based) depending on how that improved of the cross-validated log-likelihood. 

\begin{table}[h]
	\caption{ Reviewed articles about Bayesian networks, KDEs and HMMs }
	\centering
	\begin{tabular}{lll}
		\textbf{Area }&\textbf{Name} & \textbf{Short summary} \\
		\hline
		\textbf{BNs}  &  \cite{1995:hofmann.tresp}                & Introduced KCE in BN            \\
		&  \cite{2009:perez.larranaga.ea}           & KCE in Bayesian classifiers     \\
		&  \cite{2022:puerto-santana.larrnaga.ea}   & Context-specific LGBN in HMMs   \\
		&  \cite{2022:atienza.bielza.ea}            & LG or KCE for BN structure      \\  
		\hline
		\textbf{HMMs} &	 \cite{1990:rajarshi}                     & Markov process with KDE         \\
		&  \cite{2003:wang.zheng.ea}                & Input-output KDE in HMMs        \\ 
		&	 \cite{2005:xu.wu.ea}                     & AR values as kernel centres     \\
		& 									      & with KDE emission probabilities \\
		&	 \cite{2007:piccardi.perez}               & KDE-HMMs treated  as MoG-HMMs   \\
		&	 \cite{2014:do.xiao.ea}                   & KDE in MoGs/HMMs with ANNs      \\
		&	 \cite{2017:qiao.xi}                      & KDE in Markov random fields     \\
		&											  & with Gibbs sampling             \\	
		&	 \cite{2018:henter.leijon.ea}             & AR-HMM conditional emission     \\
		&                                           & distribution defined by KDE     \\
		&	 \cite{2021:luati.novelli}                & KDE in transitions in HSMMs     \\
		&	 \cite{2022:jung.park}                    & KDE in the covariance in HMMs   \\
		&											  & with Gaussian processes         \\  			
		&  This article                    & HMMs with KDE emissions         \\
		&											  & with context-specific BN        \\
	\end{tabular}
	
	\label{ table:sota_offline }
\end{table}

\subsection{Hidden Markov models}

Regarding HMMs,  \cite{1990:rajarshi}  proposed a bootstrap method based on KDE to define stationary Markov processes. 
Although no hidden variable was used in this first work, these ideas were generalized to be adapted in HMMs, as it will be seen below.
\cite{2003:wang.zheng.ea} proposed an HMM with input-output observations.
The model assumed that in each hidden state, the output variables would depend on the input variables in a manner described by a BN, and the conditional probabilities were estimated using KCE.
The authors proposed an EM algorithm with a Monte Carlo sampling phase in the M-step to reduce the number of instances in the kernel and accelerate the training phase.
\cite{2005:xu.wu.ea} proposed a kernel-based HMM, where the emission probabilities were a KDE model whose kernel took as arguments the current and last observations.
The model learning procedure was modified to maximize the accuracy in a supervised problem.
Another approach to join HMMs and KDEs was proposed in \cite{2007:piccardi.perez}, where an HMM with kernel-based emission probability estimation was proposed, but, in this case, a pseudo-likelihood function was used to run the EM algorithm.
For the multivariate case, the bandwidths were defined with a matrix in order to take into consideration feature interactions. 
\cite{2014:do.xiao.ea} proposed a kernel-based density HMM for classification  in speech recognition.
The emission probabilities were based on KDEs; however, in this case, a global bandwidth parameter was used.
Next, an artificial neural network (ANN) was used to re-estimate the a-posteriori probabilities of the cluster/class variable to improve the accuracy of the speech recognition.

In more recent years, more work related to HMMs and KDEs can be found.
\cite{2017:qiao.xi} proposed a kernel-based hidden Markov random field model, where the emission probabilities were modeled as KDEs; but the latent probabilities were modeled with a Gibbs distribution.
An EM algorithm was proposed to determine the model parameters, in particular, the bandwidth of the kernels depended on the hidden state. 
\cite{2018:henter.leijon.ea}, proposed an AR-HMM where a KDE was used to define the conditional next-step emission distributions.
An extended EM algorithm was applied to learn the model parameters; also, the bandwidths and kernel centers were dependent on each hidden state.
This was subsequently extended to kernel conditional density estimation by \cite{2022:de-gooijer.henter.ea}.
In \cite{2021:luati.novelli}, the KDEs were used to estimate the sojourn times distributions for explicit state duration in hidden semi-Markov models (HSMM). 
Finally, \cite{2022:jung.park} proposed an HMM with Gaussian processes as emission probabilities.
For the covariance function, a spectral mixture kernel was applied, and the model parameters were learned using variational Bayesian methods.

In a final related publication, \cite{2022:puerto-santana.larrnaga.ea} proposed a context-specific linear Gaussian HMM, called AR-AsLG-HMM.
They showed the model to be capable of estimating the AR order and BN dependence structure for each separate variable, depending on each hidden state
KDE was not applied, limited the applications to Gaussian data.
Asymmetric models have previously been proposed by \cite{2003:bilmes}, \cite{2004:kirshner.padhraic.ea} and \cite{2017:bueno.hommersom.ea}, but none of them treat the issue of modeling non-Gaussian data for continuous variables.
Our proposed model mixes both KDEs and As-HMMs using context-specific BNs, allowing the emission probabilities to learn arbitrary data distributions, and share information between variables.  

From the reviewed articles, we observe that there is no HMM model with KDE emission probabilities which uses context-specific BNs to express variable dependencies. 
Such expressions can reduce the number of parameters and prevent overfitting \cite{2017:bueno.hommersom.ea}.
Additionally, the use of such context-specific BNs can provide data insights and describe the relationships between variables.
On the other hand, current asymmetric models rely on the Gaussian distribution, which can limit their expressiveness to represent more general data.
In this article, we address these issues introducing a model that uses KDE to model emission probabilities and uses context-specific BNs to share information between variables when necessary. 

\section{Theoretical background}\label{sec:background}

In this section, a summary of the relevant background knowledge is provided.
Since the proposal is related to HMMs and KDEs, these models are reviewed first.
Additionally, the structural EM algorithm is explained since it will be used for structure learning in the proposed KDE-AsHMMs. 

\subsection{Kernel density estimation}\label{sec2:kernel_fund}

When data is being analyzed, histograms are the first approximation to estimate the underlying data distribution in a nonparametric manner. 
The number, width and position of the bins must be found by trial and error and the final estimation is not continuous nor smooth as mentioned in \cite{1986:silverman}.
A more sophisticated and general approximation was introduced in \cite{1956:rosenblatt} and \cite{1962:parzen}, where the density was estimated using a mixture of uniform distributions, weighted by a bandwidth $h$ (similar to the bin width in a histogram).
This approach was subsequently generalized in the following manner: let $K:\mathbb{R} \rightarrow \mathbb{R}$  be a symmetric probability density function, also known as the kernel, $h\in\mathbb{R}$ a bandwidth, $\mathcal{D}= \{y^0,...,y^{T-1}\}$ training i.i.d. samples drawn from a random variable $Y$  with unknown distribution.
A kernel density estimation (KDE) of the probability density function $f_Y(y)$ is defined as:
\begin{equation}
	\hat{f}_Y(y|\mathcal{D},h) = \sum_{t=0}^{T-1}\frac{1}{Th}K\left(\frac{y-y^t}{h}\right).
\end{equation} 

There are several design choices to consider when working with kernels, such as the selection of the kernel function $K$ and the bandwidth $h$.
\cite{1962:parzen} points out that the selection for $K$ is not crucial in terms of asymptotic behavior regarding the approximate mean integral squared error (AMISE).
In \cite{1986:silverman}, rule of thumb  bandwidth settings can be found such that $h$ minimizes the AMISE.


\subsection{Hidden Markov models}
Let $\boldsymbol{X}^{0:T}= (\boldsymbol{X}^0,...,\boldsymbol{X}^T)$ be an observable stochastic process with $\boldsymbol{X}^t=(X^t_1,...,X^t_M)$ a random vector of $M$ variables/features. 
Assume that the process $\boldsymbol{X}^{0:T}$ relies on a hidden or non-observable stochastic process $\boldsymbol{Q}^{0:T}= (Q^0,...,Q^T)$, where the  values of the range of $Q^t$ is finite, i.e, $R(Q^t) :=\{1,2,...,N\}$, $t=0,1,...,T$.
These values are called states and determine the process $\boldsymbol{X}^{0:T}$. 
A hidden Markov model (HMM) is a double chain stochastic process, where the stochastic hidden process $\boldsymbol{Q}^{0:T}$ is assumed to satisfy the Markov property, i.e., $P(Q^t|\boldsymbol{Q}^{0:t-1}) = P(Q^t|Q^{t-1})$. 
$\boldsymbol{X}^{0:T}$ is usually assumed to be independent of itself over time and dependent on $\boldsymbol{Q}^{0:T}$ i.e., $P(\boldsymbol{X}^t|\boldsymbol{X}^{0:t-1},\boldsymbol{Q}^{0:t}) = P(\boldsymbol{X}^t|Q^t)$.

In a more formal way, an HMM can be defined  as a triplet $\boldsymbol{\lambda}=(\textbf{A},\textbf{B},\boldsymbol{\pi})$ with $\textbf{A}= [a_{ij}]_{i,j=1}^N$, where $a_{ij}=P(Q^{t+1}=j|Q^t=i)$.
$\textbf{B} = [b_j(\boldsymbol{X}^t)]_{j=1}^N$ is a vector representing the emission probability of the observations given the hidden state; if $\boldsymbol{X}^t$ is discrete and its range has $\kappa$ possible values, then  $b_j(\boldsymbol{X}^t)=[P(\boldsymbol{X}=k|Q^t=j)]_{k=1}^{\kappa}$ and $\textbf{B}$ has dimension $\kappa\times N$.
If $\boldsymbol{X}^t$ is continuous, $b_j(\boldsymbol{X}^t)=f(\boldsymbol{X}^t|Q^t=j)$ and the components of $\textbf{B}$ are probability density functions.
Finally, $\boldsymbol{\pi}$ is the initial distribution of hidden states:  $\boldsymbol{\pi}=[\pi_j]_{j=1}^N$, where $\pi_j = P(Q^0= j)$.

Three main tasks can be performed in the context of HMMs: first, given a model $\boldsymbol{\lambda}$, compute the likelihood of a new instance $\boldsymbol{x}$, i.e., $P(\boldsymbol{x}|\boldsymbol{\lambda})$, which can be done using the forward step from the forward-backward algorithm. 
Second, given a model $\boldsymbol{\lambda}$, estimate the most probable sequence of hidden states for a set of observations $\boldsymbol{x}^{0:L}$ which can be solved using the Viterbi algorithm. 
Third, learn the parameter $\boldsymbol{\lambda}$, which is usually  estimated with the EM algorithm or Baum-Welch algorithm for the specific case of HMMs, see  \cite{1977:dempster.laird.ea}, to approximate the maximum likelihood estimators.
These algorithms are further detailed in \cite{1990:rabiner}.

\subsection{Structural expectation maximization algorithm}

In \cite{1998:friedman} the structural EM (SEM) algorithm was introduced  as a generalization of the EM algorithm.
The SEM algorithm was developed to jointly estimate the parameters $\boldsymbol{\lambda}$ and structure $\mathcal{B}$ of a model.
In this sense, SEM helps us to find the desired model structure and parameters in the presence of arbitrary hidden variables $\boldsymbol{H}$ (in our case $\boldsymbol{Q})$.
To protect against overfitting, the SEM algorithm includes a penalty term similar to the Bayesian information criterion (BIC)  (see \cite{1978:schwarz})  that  depends on $\#(\mathcal{B})$, the number of parameters of the model structure.
In this manner, the networks $\mathcal{B}$ are expected to be enough complex to explain the data preventing overtfitting. 
Assume that a structure with parameters $\mathcal{B}^{(s)}$, $\boldsymbol{\lambda}^{(s)}$ has already been computed.
The algorithm iteratively optimizes the auxiliary function $\mathcal{Q}(\mathcal{B}, \boldsymbol{\lambda}|\mathcal{B}^{(s)},\boldsymbol{\lambda}^{(s)})$ to discover the model structure and parameters.
The auxiliary function has the form::
\begin{equation}\label{eq:Q_aux_sem}
	\mathcal{Q}(\mathcal{B},\boldsymbol{\lambda}|\mathcal{B}^{(s)},\boldsymbol{\lambda}^{(s)}) = \mathbb{E}_{P(\boldsymbol{H}|\boldsymbol{x},\mathcal{B}^{(s)},\boldsymbol{\lambda}^{(s)})}[\ln f(\boldsymbol{x},\boldsymbol{h}|\mathcal{B},\boldsymbol{\lambda})]-0.5\#(\mathcal{B})\ln(T+1).
\end{equation}

An iteration of the algorithm proceeds as follows:
\begin{itemize}
	\item[1.] Use a structure search algorithm\footnote{The structure search algorithm can be seen as a combinatorial optimization problem. 
		The problem can be solved using heuristics like hill-climbing or meta-heuristics, like tabu search, see \cite{1986:glover}, simulated annealing, see \cite{1983:kirkpatrick.gelatt.ea}, or genetic algorithm, see \cite{1998:mitchell}, and so on.}  to look for candidate structures.
	\item[2.] For every candidate structure, use the current a-posteriori latent probabilities to perform the M-step, and estimate the parameters of the structure.
	\item[3.] For every candidate structure, use Eq.~\ref{eq:Q_aux_sem} to compute its scores.
	\item[4.] Set $\mathcal{B}^{(s+1)}$ equal to the candidate structure that maximizes the score.
	\item[5.] set $\boldsymbol{\lambda}^{(s+1)}$ equal to the parameter estimates found by the EM algorithm.
\end{itemize}
Steps 1 through 5 are iterated until a stopping criterion is satisfied, for example the increase in score is below a threshold value or a certain maximum number of iterations is reached.

\section{Proposed model}\label{sec:proposed}

\subsection{Definition}

\begin{figure}[h]
	\centering
	\begin{tikzpicture} 
		[->,thick, scale=0.8,auto=center, every node/.style={scale=0.7}]
		
		\node (q1)[style={circle,fill=black!35,draw=black!100}] at (0,2.25)  {$Q^{t+1}$};
		\node (q2)[style={circle,fill=black!35,draw=black!100}] at (2.5,2.25)  {$Q^{t+2}$};
		\node (q3)[style={circle,fill=black!35,draw=black!100}] at (5,2.25)  {$Q^{t+3}$};
		\node (q4)[style={circle,fill=black!35,draw=black!100}] at (7.5,2.25)  {$Q^{t+4}$};
		
		\node (x111)[style={circle,fill=gray!20,draw=black!100}] at (0,-0.75)  {$X_1^{t+1}$};
		\node (x121)[style={circle,fill=gray!20,draw=black!100}] at (2.5,-0.75)  {$X_1^{t+2}$};
		\node (x131)[style={circle,fill=gray!20,draw=black!100}] at (5,-0.75)  {$X_1^{t+3}$};
		\node (x141)[style={circle,fill=gray!20,draw=black!100}] at (7.5,-0.75)  {$X_1^{t+4}$};
		
		\node (x112)[style={circle,fill=gray!20,draw=black!100}] at (0,-2.25)  {$X_2^{t+1}$};
		\node (x122)[style={circle,fill=gray!20,draw=black!100}] at (2.5,-2.25)  {$X_2^{t+2}$};
		\node (x132)[style={circle,fill=gray!20,draw=black!100}] at (5,-2.25)  {$X_2^{t+3}$};
		\node (x142)[style={circle,fill=gray!20,draw=black!100}] at (7.5,-2.25)  {$X_2^{t+4}$};
		
		\node (z11)[style={circle,fill=black!35,draw=black!100}] at (-1,1)  {$\boldsymbol{W}^{t+1}$};
		\node (z12)[style={circle,fill=black!35,draw=black!100}] at (1.5,1)  {$\boldsymbol{W}^{t+2}$};
		\node (z13)[style={circle,fill=black!35,draw=black!100}] at (4,1)  {$\boldsymbol{W}^{t+3}$};
		\node (z14)[style={circle,fill=black!35,draw=black!100}] at (6.5,1)  {$\boldsymbol{W}^{t+4}$};
		
		\node (x212)[style={circle,fill=gray!20,draw=black!100}] at (0,5.25)  {$X_2^{t+1}$};
		\node (x222)[style={circle,fill=gray!20,draw=black!100}] at (2.5,5.25)  {$X_2^{t+2}$};
		\node (x232)[style={circle,fill=gray!20,draw=black!100}] at (5,5.25)  {$X_2^{t+3}$};
		\node (x242)[style={circle,fill=gray!20,draw=black!100}] at (7.5,5.25)  {$X_2^{t+4}$};
		
		\node (x211)[style={circle,fill=gray!20,draw=black!100}] at (0,6.75)  {$X_1^{t+1}$};
		\node (x221)[style={circle,fill=gray!20,draw=black!100}] at (2.5,6.75)  {$X_1^{t+2}$};
		\node (x231)[style={circle,fill=gray!20,draw=black!100}] at (5,6.75)  {$X_1^{t+3}$};
		\node (x241)[style={circle,fill=gray!20,draw=black!100}] at (7.5,6.75)  {$X_1^{t+4}$};
		
		\node (z21)[style={circle,fill=black!35,draw=black!100}] at (-1,3.5)  {$\boldsymbol{W}^{t+1}$};
		\node (z22)[style={circle,fill=black!35,draw=black!100}] at (1.5,3.5)  {$\boldsymbol{W}^{t+2}$};
		\node (z23)[style={circle,fill=black!35,draw=black!100}] at (4,3.5)  {$\boldsymbol{W}^{t+3}$};
		\node (z24)[style={circle,fill=black!35,draw=black!100}] at (6.5,3.5)  {$\boldsymbol{W}^{t+4}$};
		
		\draw [black] (0,-1.5) ellipse (-0.75 and -1.75)  ;
		\draw [black] (2.5,-1.5) ellipse (-0.75 and -1.75) ;
		\draw [black] (5,-1.5) ellipse (-0.75 and -1.75) ;
		\draw [black] (7.5,-1.5) ellipse (-0.75 and -1.75) ;
		
		\draw (x211) -> (x212);
		\draw (x221) -> (x222);
		\draw (x231) -> (x232);
		\draw (x241) -> (x242);
		
		\draw (-1.5,-0.75) -> (x111); 
		\draw (x111) -> (x121);
		\draw (x121) -> (x131);
		\draw (x131) -> (x141);
		\draw (x141) -> (8.5,-0.75);
		\draw (-1.5,-2.25) -> (x112); 
		\draw (x112) -> (x122);
		\draw (x122) -> (x132);
		\draw (x132) -> (x142);
		\draw (x142) -> (8.5,-2.25);
		\draw (-1.5,-3.25) to[out=0,in=-135] (x112);
		\draw (x112) to[out=-45,in=-135] (x132);
		\draw (x122) to[out=-45,in=-135] (x142);
		\draw (-1.5,-2.75) to[out=-45, in=-135] (x122);
		\draw (x132) to[out=-45,in=-135] (8.5,-2.75);
		\draw (x142) to[out=-45,in=180] (8.5,-3.);
		
		\draw (q1) -> (q2);
		\draw (q2) -> (q3);
		\draw (q3) -> (q4);
		\draw (-1.5,2.25) -> (q1);
		\draw (q4) -> (8.5,2.25);
		
		\draw (q1) -> node[below,rotate=90] {\footnotesize{$Q^{t+1}=2$}} (0,0.25);
		\draw (q2) -> node[below,rotate=90,] {\footnotesize{$Q^{t+2}=2$}}(2.5,0.25);
		\draw (q3) -> node[below,rotate=90] {\footnotesize{$Q^{t+3}=2$}}(5,0.25);
		\draw (q4) -> node[below,rotate=90] {\footnotesize{$Q^{t+4}=2$}}(7.5,0.25);
		
		\draw (z11) -> (-0.2,0.23);
		\draw (z12) -> (2.3,0.23);
		\draw (z13) -> (4.8,0.23);
		\draw (z14) -> (7.3,0.23);
		
		\draw  (q1.south west) -- ( z11.north east);
		\draw  (q2.south west) --( z12.north east);
		\draw  (q3.south west) --( z13.north east);
		\draw  (q4.south west) --( z14.north east);
		
		\draw  (q1.north west) --( z21.south east);
		\draw  (q2.north west) --( z22.south east);
		\draw  (q3.north west) --( z23.south east);
		\draw  (q4.north west) --( z24.south east);
		
		\draw [black] (0,6) ellipse (-0.75 and -1.75) ;
		\draw [black] (2.5,6) ellipse (-0.75 and -1.75) ;
		\draw [black] (5,6) ellipse (-0.75 and -1.75) ;
		\draw [black] (7.5,6) ellipse (-0.75 and -1.75) ;

		\draw (z21) -> (-0.2,4.27);
		\draw (z22) -> (2.3,4.27);
		\draw (z23) -> (4.8,4.27);
		\draw (z24) -> (7.3,4.27);
		
		\draw (q1) -> node[below,rotate=90] {\footnotesize{$Q^{t+1}=1$}} (0,4.25);
		\draw (q2) -> node[below,rotate=90] {\footnotesize{$Q^{t+2}=1$}}(2.5,4.25);
		\draw (q3) -> node[below,rotate=90] {\footnotesize{$Q^{t+3}=1$}}(5,4.25);
		\draw (q4) -> node[below,rotate=90] {\footnotesize{$Q^{t+4}=1$}}(7.5,4.25);
		
	\end{tikzpicture}
	\caption{Example of a KDE-AsHMM pictured as a dynamic BN}
	\label{fig:kdehmmgraph}
\end{figure}
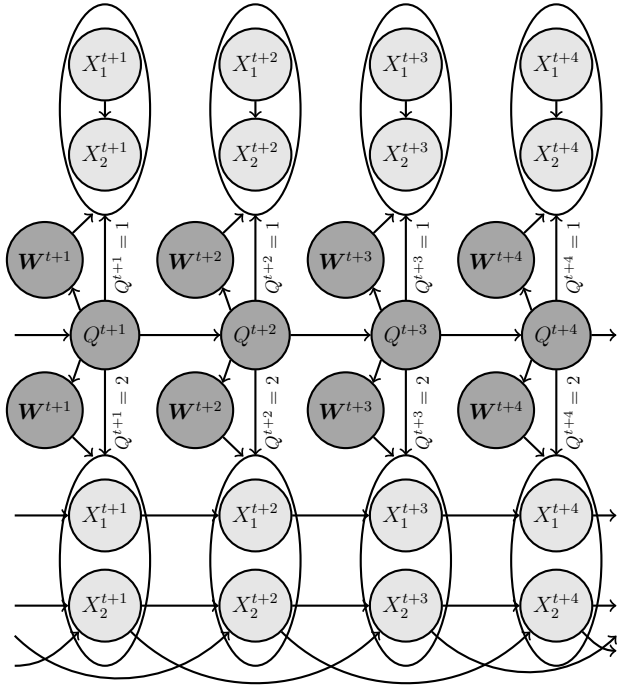
The core idea behind this new model is to describe and identify non-Gaussian and non-stationary dynamic processes by combining HMMs and KDEs. 
The proposed model, denoted KDE-AsHMM, can be described as a dynamic BN, see Figure~\ref{fig:kdehmmgraph}.
In the figure, the added latent variables $\boldsymbol{W}^t$ follow a categorical distribution, which depends on the latent variables $\boldsymbol{Q}^t$.
The latent variable $\boldsymbol{W}^t$ is used to determine the most representative instances to be used in the kernel density estimation for each hidden state.
The observable variables $\boldsymbol{X}^t$, can depend statistically on each other through a directed acyclic graph structure, as well autoregressive dependences on their most recent prior values, up to a maximum order $P*$.
This value $P^*$ can be chosen by the experimenter or calculated using the Box-Jenkins methodology, see \cite{1976:box.jenkins}. 
Nevertheless, these relationships may change depending on the hidden state as pictured in Figure~\ref{fig:kdehmmgraph}: when $Q^t= 1$,  $X^t_2$ depends on $X^t_1$, but when $Q^t=2$, the relationship changes and $X^t_1$ instead depends on $X^{t-1}_1$, whereas $X^t_2$ depends on $X^{t-1}_2$ and $X^{t-2}_2$.
In this manner, the complexity of the emission probability distributions can increase as the data requires.

The different conditional dependencies provided by  Figure~\ref{fig:kdehmmgraph} are defined as follows.
Assume that $N$ hidden states, $M$ variables are being modeled and the length of the training data is $L+1$ and the length of the test data is $T+1$.
Regarding the transition probabilities:
\begin{equation}
	P(Q^t=q^t|Q^{t-1}=q^{t-1};\boldsymbol{\lambda}) = a_{q^{t-1}q^t}, \text{ with }  \sum_{j=1}^Na_{q^{t-1}j} = 1.
\end{equation}
With respect to the latent variable $\boldsymbol{W}^{t} = \{W^t_{P^*}, W^t_{P^*+1},...,W^t_{L}\}$, which determines the relevant training samples for each hidden state, its conditional probabilities are defined as:
\begin{equation}\label{eq:ws}
	P(\boldsymbol{W}^t|Q^{t};\boldsymbol{\lambda}) = \prod_{l=P^*}^L(\omega_{Q^t,l})^{W^t_l}, \quad \sum_{l=P^*}^{L}\omega_{Q^tl} = 1, \quad \sum_{l=P^*}^LW_l^t =1, \quad W^t_l\in\{0,1\}.
\end{equation}

Assume that the training data  is $\boldsymbol{y}^{0:L}$.
During the training phase $T=L$, it is necessary that $W^l_l =0$ for $l=0,...,L$; in this manner, issues with degenerate likelihoods and infinitely narrow kernels are avoided, see \cite{2007:piccardi.perez}. 

Let $\text{Set}(\boldsymbol{X}^t) := \{X_1^t,...,X_M^t\}$ the set of features of the model.
We denote the $\kappa_{q^tm}$ parents of $X_m^t$ at $q^t\in R(Q^t)$ as $\{V^t_{q^tmk}\}_{k=1}^{\kappa_{q^tm}}\subset\text{Set}(\boldsymbol{X}^t)$.
With this notation, the conditional densities of the observable variables in this article are defined as:
\begin{equation}\label{eq:f_w}
	f_{\boldsymbol{X^t}|\boldsymbol{W}^t,Q^t}(\boldsymbol{X}^t|\boldsymbol{W}^t,Q^t;\boldsymbol{\lambda}) = \prod_{m=1}^M	f_{X_m^t|\boldsymbol{W}^t,Q^t,\boldsymbol{U}^t_{Q^t,m}}(X_m^t|\boldsymbol{W}^t,Q^t,\boldsymbol{U}^t_{Q^t,m};\boldsymbol{\lambda}),
\end{equation}
where each $\boldsymbol{U}^t_{Q^tm}= (V^t_{Q^tm1},...,V^t_{im\kappa_{Q^tm}},X^{t-1}_m,...,X^{t-p_{Q^tm}}_m)$ is a context-specific random vector, which contains the $\kappa_{Q^tm}$ parents and $p_{Q^tm}$ AR dependencies of the variables $X^t_m$.
In order to have an interpretable and well defined model, for every $q^t\in R(Q^t)$, the set of tuples $\bigcup_{m=1}^{M}\boldsymbol{U}^t_{q^tm}\times\{X_m\}$, must describe the edges of an directed acyclic graph (DAG), i.e., a context-specific BN\footnote{Here, $\times$ represents the Cartesian product}.

To analytically describe, the dependencies in the context-specific BN,
define $\boldsymbol{v}^l_{Q^tm}$ as the sample value of the random vector $\boldsymbol{U}^t_{Q^t,m}$ by $\boldsymbol{y}^{0:L}$, and define $\boldsymbol{M}_{Q^t,m}$ as a context-specific matrix of size $(\kappa_{Q^tm}+p_{Q^tm})\times(\kappa_{Q^tm}+p_{Q^tm})$ which determines the weights of the kernel dependencies, and set:
\begin{equation}\label{eq:moscato}
	\mu_{l,Q^t,m}^t := y_m^l +\boldsymbol{M}_{Q^t,m}(\boldsymbol{U}^t_{Q^t,m}-\boldsymbol{v}^l_{Q^t,m})^{\top}.
\end{equation}
We rewrite each factor in Eq.~(\ref{eq:f_w}) in terms of a kernel function $K$, bandwidth $h_{Q^t,m}$ and centers $\mu_{l,Q^t,m}^t$:
\begin{equation}
	f_{X_m^t|\boldsymbol{W}^t,Q^t,\boldsymbol{U}^t_{Q^t,m}}(X_m^t|\boldsymbol{W}^t,Q^t,\boldsymbol{U}^t_{Q^t,m};\boldsymbol{\lambda}) := \prod_{l=P^*}^{L}\left(\frac{1}{h_{Q^t,m}}K\left( \frac{X^t_m -\mu_{l,Q^t,m}^t}{h_{Q^t,m}} \right)\right)^{W^t_l}.
\end{equation}
The previous equations omit the dependencies on  $\boldsymbol{y}^{0:L}$ and $\boldsymbol{X}^{t-P^*:t-1}$ to simplify notation.
Note additionally that the bandwidths are allowed to vary for each variable for each hidden state.
In this manner, the deviations on parents and AR values can be used to correct the kernel as the data requires. 
Notice that every component of the KDE  for each hidden state can be obtained  using Eq.~(\ref{eq:ws})--(\ref{eq:moscato}) as follows: 
\begin{equation}
	f_{\boldsymbol{X}^t,W^t_l=1|Q^t=i}(\boldsymbol{X}^t,W^t_l=1|Q^t=i;\boldsymbol{\lambda}) = \omega_{il}\prod_{m=1}^M\frac{1}{h_{im}}K\left( \frac{X^t_m -\mu_{lim}^t}{h_{im}} \right).
\end{equation}
This is useful to determine the emission probabilities for this model:
\begin{equation}
	b_i(\boldsymbol{X}^t) = \sum_{l= P^*}^T \omega_{il}\prod_{m=1}^M\frac{1}{h_{im}}K\left( \frac{X^t_m -\mu_{lim}^t}{h_{im}} \right).
\end{equation}

The full information would be:
\begin{equation}
	\begin{aligned}
		f&_{\boldsymbol{X}^{P^*:T},\boldsymbol{Q}^{P^*:T},\boldsymbol{W}^{P^*:T}}(\boldsymbol{x}^{P^*:T},\boldsymbol{q}^{P^*:T},\boldsymbol{w}^{P^*:T};\boldsymbol{\lambda}) = \\ &\pi_{q^{P^*}}\prod_{t=P^*}^{T-1}a_{q^t,q^{t+1}}\prod_{t=P^*}^{T}\prod_{l=P^*}^L\left(\omega_{q^t,l}\prod_{m=1}^M\frac{1}{h_{q^t,m}}K\left( \frac{X^t_m -\mu_{l,q^t,m}^t}{h_{q^t,m}} \right)\right)^{w^t_l}
	\end{aligned}
\end{equation}
In log form 
\begin{equation}
	\begin{aligned}
		\ln(f) =& \ln(\pi_{q^{P^*}})+\sum_{t=P^*}^{T-1}\ln(a_{q^t,q^{t+1}})+\\
		&\sum_{t= P^*}^{T}\sum_{l=P^*}^Tw^t_l\left(\ln(\omega_{q^t,l})+\sum_{m=1}^M\ln\left(\frac{1}{h_{q^t,m}}K\left( \frac{X^t_m -\mu_{l,q^t,m}^t}{h_{q^t,m}}\right)\right)\right)
	\end{aligned}
\end{equation}

As summary, the proposed model $\boldsymbol{\lambda}$ consists of the set of  parameters $ \boldsymbol{\lambda}~:=~\{\boldsymbol{\pi}, \textbf{A}, \boldsymbol{h} := \{h_{im}\}_{i=1,m=1}^{N,M}, \boldsymbol{\omega} := \{\omega_{il}\}_{i=1,l=P^*}^{N,L}, \boldsymbol{M} := \{\boldsymbol{M}_{im}\}_{i=1,m=1}^{N,M} \}$, where $\boldsymbol{M}_{im}$ represents the  $\kappa_{im}$ dependencies of $X_m$ from other variables and its $p_{im}\leq P^*$ AR dependencies. 
From the indexing, the dependencies can vary with $X_m$ and the hidden state $i\in R(Q^t)$.
Also, the dependencies must follow a BN, and therefore DAG structures must be employed. 

\subsection{Learning algorithm}
For this model, the EM algorithm will be applied to learn the models parameters, assume that a model $\boldsymbol{\lambda}^{(s)}$ has been computed or provided.
The auxiliary function for this model is:
\begin{equation}\label{eq:auxkde}
	\begin{aligned}
		\mathcal{Q}&(\boldsymbol{\lambda}|\boldsymbol{\lambda}^{(s)}) = \\ &E_{P(\boldsymbol{Q}^{P^*:T},\boldsymbol{W}^{P^*:T}|\boldsymbol{X}^{0:T};\boldsymbol{\lambda}^{(s)})}[\ln f_{\boldsymbol{X}^{P^*:T},\boldsymbol{Q}^{P^*:T},\boldsymbol{W}^{P^*:T}}(\boldsymbol{x}^{P^*:T},\boldsymbol{Q}^{P^*:T},\boldsymbol{W}^{P^*:T};\boldsymbol{\lambda})]
	\end{aligned}
\end{equation}
Recall that for the training phase $\boldsymbol{x}^{0:T} = \boldsymbol{y}^{0:L}$ and $W^l_{l} =0$, and thence Eq.~(\ref{eq:auxkde}) can be expressed analytically as:
\begin{equation} \label{eq:Qkde}
	\begin{aligned}
		\mathcal{Q}(\boldsymbol{\lambda}|\boldsymbol{\lambda}^{(s)}) &= \sum_{i=1}^N\gamma^0(i)\ln(\pi_{i})+\sum_{t=P^*}^{T-1}\sum_{i=1}^N\sum_{j=1}^N\zeta^t(i,j)\ln(a_{ij})+\\
		&\sum_{t=P^*}^{T}\sum_{l\not=t}^T\sum_{i=1}^N\psi_l^t(i)\left(\ln(\omega_{il})+\sum_{m=1}^M\ln\left(\frac{1}{h_{im}}K\left( \frac{x^t_m -\mu_{lim}^t}{h_{im}}\right)\right)\right)
	\end{aligned}
\end{equation}

From the previous equation, the  following latent variables appear:

\begin{equation}
	\begin{aligned}
		\gamma^t(i) &= P(Q^t=i|\boldsymbol{x}^{0:T};\boldsymbol{\lambda}^{(s)} ) \\
		\zeta^t(i,j) &= P(Q^t=i,Q^{t+1}=j|\boldsymbol{x}^{0:T};\boldsymbol{\lambda}^{(s)}) \\
		\psi^t_l(i) &= P(Q^t=i,W^t_l=1|\boldsymbol{x}^{0:T};\boldsymbol{\lambda}^{(s)})
	\end{aligned}
\end{equation}

In particular, $\psi^t_l(i)$ can be computed as follows:


\begin{equation}
	\psi_l^t(i) = \frac{\omega_{il}\prod_{m=1}^MK\left( \frac{x^t_m -\mu_{lim}^t}{h_{im}} \right)}{\sum_{k\not=t}^T\omega_{ik}\prod_{m=1}^MK\left( \frac{x^t_m -\mu_{kim}^t}{h_{im}} \right)}\gamma^t(i).
\end{equation}

On the other hand, $\gamma^t(i)$  and $\zeta^t(i,j)$ can be computed using the well known forward-backward algorithm.
Let us assume that we have a Gaussian kernel:
\begin{equation}
	K(x) = \frac{1}{\sqrt{2\pi}}e^{-\frac{x^2}{2}}
\end{equation}

Set $\overline{\boldsymbol{u}}^t_{iml} := (\boldsymbol{u}^t_{im}-\boldsymbol{v}^l_{im}) $ and $\overline{x}^t_{ml} := (x^t_m-y^l_m)$. 
For the M step, the updating formulas for the parameters $\boldsymbol{M}_{im}$ are deduced as:

\begin{equation}\label{eq:upm1}
	\frac{\partial\mathcal{Q}(\boldsymbol{\lambda}|\boldsymbol{\lambda}^{(s)})}{\partial\boldsymbol{M}_{im}} = \sum_{t=P^*}^T\sum_{l\not=t}^T\psi_l^t(i)\frac{\partial}{\partial \boldsymbol{M}_{im}}\left(\ln\left(K\left(\frac{x^t_m-\mu^t_{lim}}{h_{im}}\right)\right)\right)  =0
\end{equation}
\begin{equation}\label{eq:upm4}
	\left(\sum_{t=P^*}^T\sum_{l\not=t}^T\psi_l^t(i)(\overline{\boldsymbol{u}}^t_{iml})^\top\overline{\boldsymbol{u}}^t_{iml}\right)\boldsymbol{M}_{im}^\top = \sum_{t=P^*}^T\sum_{l\not=t}^T\psi_l^t(i)(\overline{\boldsymbol{u}}^t_{iml})^\top\overline{x}^t_{ml} 
\end{equation}
\begin{equation}\label{eq:upm5}
	(\boldsymbol{M}_{im}^{\top})^{(s+1)} = \left(\sum_{t=P^*}^T\sum_{l\not=t}^T\psi^t_l(i)(\overline{\boldsymbol{u}}^t_{iml})^{\top}\overline{\boldsymbol{u}}^t_{iml}\right)^{-1}\left(\sum_{t=P^*}^T\sum_{l\not=t}^T\psi^t_l(i)(\overline{\boldsymbol{u}}^t_{iml})^{\top}\overline{x}^t_{ml}\right)
\end{equation}

In practice, the linear system in Eq.~(\ref{eq:upm4}) is solved without computing inverse matrices, in this manner, computational problems such as numerical underflow/overflow are avoided.
To see that the previous update formula corresponds to a local-maximum, note that the second derivative of Eq.~(\ref{eq:Qkde}) with respect to $\boldsymbol{M}_{im}$ is:

\begin{equation}
	\frac{\partial^2\mathcal{Q}(\boldsymbol{\lambda}|\boldsymbol{\lambda}^{(s)})}{\partial\boldsymbol{M}_{im}^2} = -\left(\sum_{t=P^*}^T\sum_{l\not=t}^T\psi^t_l(i)(\overline{\boldsymbol{u}}^t_{iml})^{\top}\overline{\boldsymbol{u}}^t_{iml}\right)
\end{equation}

Which is a weighed sum of covariance matrices, which, due to the negative sign, results in a negative-semidefinite matrix, and therefore a local-maximum. 
If $\hat{\mu}^t_{lim} = x^l_m + \boldsymbol{M}_{im}^{(s+1)}(\overline{\boldsymbol{u}}^t_{iml})^{\top}$, the updating formula for $h_{im}$ is:

\begin{equation}\label{eq:uph1}
	\frac{\partial\mathcal{Q}(\boldsymbol{\lambda}|\boldsymbol{\lambda}^{(s)})}{\partial h_{im}} = \sum_{t=P^*}^T\sum_{l\not=t}^T\psi_l^t(i)\frac{\partial}{\partial h_{im}}\left(\ln\left(\frac{1}{h_{im}}K\left(\frac{x^t_m-\hat{\mu}^t_{lim}}{h_{im}}\right)\right)\right)  =0
\end{equation}
\begin{equation}\label{eq:uph4}
	\sum_{t=P^*}^T\gamma^t(i)\frac{1}{h_{im}} = \sum_{t=P^*}^T\sum_{l\not=t}^T\psi_l^t(i)\frac{\left( x^t_m-\hat{\mu}^t_{lim} \right)^2}{h_{im}^3} 
\end{equation}
\begin{equation}\label{eq:uph5}
	h_{im}^{(s+1)} = \left(\frac{\sum_{t=P^*}^T\sum_{l\not= t}^T\psi^t_l(i)(x^t_m-\hat{\mu}^t_{lim})^2}{\sum_{t=P^*}^T\gamma^t(i)}\right)^{\frac{1}{2}}
\end{equation}
In the Eq.~(\ref{eq:uph4}), we have used that $\gamma^{t}(i) = \sum_{l\not=t}^{T}\psi^t_l(i)$.
To see whether the previous parameter estimate corresponds to a local maximum, the second derivative is computed:

\begin{equation}
	\frac{\partial^2\mathcal{Q}(\boldsymbol{\lambda}|\boldsymbol{\lambda}^{(s)})}{\partial h_{im}^2} = -\sum_{t=P^*}^T\sum_{l\not=t}^T\frac{\psi_l^t(i)}{h_{im}^2}\left( \frac{3\left( x^t_m-\hat{\mu}^t_{lim} \right)^2}{h_{im}^2} -1  \right).
\end{equation}

In this case, the second derivative is not always negative or non positive.
To ensure that the second derivative is negative observe that:
\begin{equation}
	-\sum_{t=P^*}^T\sum_{l\not=t}^T\frac{\psi_l^t(i)}{h_{im}^2}\left( \frac{3\left( x^t_m-\hat{\mu}^t_{lim} \right)^2}{h_{im}^2} -1  \right)<0
\end{equation}
\begin{equation}\label{eq:conditionh}
	h_{im} < \left( 3\sum_{t=P^*}^T\sum_{l\not=t}^T\psi_l^t(i)\left( x^t_m-\hat{\mu}^t_{lim} \right)^2\right)^{\frac{1}{2}}
\end{equation}

%
In Eq.~(\ref{eq:conditionh}) it is observed that $h_{im}$ corresponds to a local-maximum, if and only if it is lower than the root of the weighted mean of the squares of the data deviations from the kernel centers (scaled by $\sqrt{3}$).
It is relevant to mention that the weights are given by the aposterioris $\psi_l^t(i)$. 
Finally, using the constraint that $\sum_{l\not=t}^T\omega_{il} =1$ and adding a Lagrange multiplier $\nu_i$, the update formula for $\omega_{il}$ is:

\begin{equation}
	\begin{aligned}\label{eq:upome}
		&\frac{\partial\mathcal{Q}(\lambda|\lambda')}{\partial \omega_{il}} = \frac{\partial}{\partial \omega_{il}}\left(\sum_{t=P^*}^T\psi_l^t(i)\ln(\omega_{il}) +\nu_i(1-\sum_{l\not=t}^T\omega_{il})\right) = 0 \\
		&\omega_{il}^{(s+1)} = \frac{\sum_{t=P^*}^T\psi_l^t(i)}{\sum_{t=P^*}^T\gamma^t(i) }
	\end{aligned}
\end{equation}
To prove that this solution correspond to a local-maximum, observe that the second derivative is less or equal to zero:
\begin{equation}
	\frac{\partial^2\mathcal{Q}(\boldsymbol{\lambda}|\boldsymbol{\lambda}^{(s)})}{\partial \omega_{il}^2}  =-\sum_{t=P^*}^T\psi_l^t(i)\frac{1}{\omega_{il}^2} \leq 0.
\end{equation}
The previous results can be summarized in the following lemma:
\begin{lemma}
	Let $W_l^l=0$ for $l=0,...,L$, and assume that during the $(s)$ iteration of the EM algorithm,  $h_{im} < \left( 3\sum_{t=P^*}^T\sum_{l\not=t}^T\psi_l^t(i)\left( x^t_m-\hat{\mu}^t_{lim} \right)^2\right)^{\frac{1}{2}}$ holds.
	The previous conditions are necessary and sufficient for the update equations in Eq.~(\ref{eq:upm5}), Eq.~(\ref{eq:uph5}) and Eq.~(\ref{eq:upome}) to be local-maximum parameters of Eq.~(\ref{eq:Qkde}), during an EM iteration.
\end{lemma}

Since the probabilistic conditions for the hidden variable $Q^t$ are not modified from the traditional HMM, the update formulas for parameters $\bf{A}$ and $\boldsymbol{\pi}$ are the same as those found in standard traditional articles such as \cite{1990:rabiner}.

\subsection{Learning the context-specific Bayesian networks}
For the SEM algorithm, we are required to optimize the following criterion
\begin{equation}\label{eq2:quijSEM}
	\begin{aligned}
		&\mathcal{Q}(\mathcal{B},\boldsymbol{\lambda}|\mathcal{B}^{(s)},\boldsymbol{\lambda}^{(s)}) = \\ &\quad E_{P\left(\boldsymbol{Q}^{P^*:T},\boldsymbol{W}^{P^*:T}|\boldsymbol{X}^{0:T};\mathcal{B}^{(s)},\boldsymbol{\lambda}^{(s)}\right)}[\ln f_{\boldsymbol{X}^{P^*:T},\boldsymbol{Q}^{P^*:T},\boldsymbol{W}^{P^*:T}}(\boldsymbol{x}^{P^*:T},\boldsymbol{Q}^{P^*:T},\boldsymbol{w}^{P^*:T};\mathcal{B},\boldsymbol{\lambda})] \\
		& \quad -0.5\#(\mathcal{B})\ln(T).
	\end{aligned}
\end{equation}

In this manner complex structures are penalized to prevent overfitting issues and and overly dense BNs.
Due to the linear nature of  Eq.~(\ref{eq:Qkde}), only the score in Eq.~(\ref{eq:scoreik}) is affected by structure modification (for a given state $Q^t=i$  and value for $X_m$, with the other terms remaining unaffected.

\begin{equation}\label{eq:scoreik}
	\text{score}_{im} = \sum_{t=P^*}^T\sum_{l\not=t}^T\psi_l^t(i)\ln\left(K\left(\frac{x^t_m-\mu^t_{lim}}{h_{im}}\right)\right)-\frac{1}{2}(\kappa_{im}+p_{im}+T+1-P^*)\ln(T).
\end{equation}
For the SEM algorithm, during the search for  new structures, the score  in Eq.~(\ref{eq:scoreik}) is then used together with the greedy-forward algorithm described in \cite{2022:puerto-santana.larrnaga.ea}.
However, we emphasize that any other heuristic or meta-heuristic can used for the model search, such as the  tabu search in \cite{2017:bueno.hommersom.ea}, or the simulated annealing of \cite{2018:puerto-santana.bielza.ea}.

\subsection{Theoretical computation time complexity}

It is well known that the methods based on kernel density estimation can be time demanding.
This section therefore provides upper-bounds from the running time of each learning algorithm in terms of big $O$ notation.
For these bounds, two scenarios are considered: One  where the BNs are dense such that no additional arc can be put into the networks, and other where all the context-specific BNs are naïve (i.e., they contain zero arcs). 
We refer to the latter bound as the \textit{light computational upper bound}.

\begin{table}[h]
	\centering 
	\begin{tabular}{lll}
		Learning algorithm step & Upper-bound & Light upper-bound  \\
		\hline
		1. Compute $b_i(\boldsymbol{x}^t)$& $O(T^2NMS)$ & $O(NT^2M)$\\
		2. Estimate $\gamma^t(i)$, $\psi^t_l(i)$& $O(TN(N^2+T))$ & $O(TN(N^2+T))$ \\
		3. Update $\textbf{A}$ and $\boldsymbol{\pi}$ & $O(NT)$ & $O(NT)$ \\
		4. Update $\boldsymbol{\omega}$ & $O(NT^2)$ & $O(NT^2)$\\
		5. Update  $\boldsymbol{M}$ & $O(NS^2(T^2+MS))$ & $O(0)$ \\
		6. Update $\boldsymbol{h}$ & $O(NMT^2)$ & $O(NMT^2)$ \\
	\end{tabular}
	\caption{Computational upper bounds.
		In the upper-bound column, it is assumed that the context-specific BNs are dense. 
		In the light upper-bound column, it is assumed that the context-specific BNs are naïve-BNs. $S = P^*+M$}
	\label{table:complexity}
\end{table}

Table~\ref{table:complexity} reports the bounds, where for the sake of space, $S:= P^*+M$.
The bounds are arranged in the same order as the respective steps in the learning algorithm.
It is noticeable that the presence of BN in the model only affects the computation of emission probabilities and the updates $\boldsymbol{M}$.
Nevertheless, the computational effort needed to update $\boldsymbol{M}$ can be high in the case of dense networks, since it is quadratic in the training input length and fourth-power in the number of variables.
The fourth-power dependency in the number of variables comes from the solution of linear systems and the loop through all the variables to solve their corresponding system. 
In this sense, it is desirable to keep the number of dependencies as low as possible; otherwise, the computational time required to train a model can be prohibitive. 

The log-likelihood of test data can be evaluated using  steps 1 and 2 from  Table~\ref{table:complexity}.
Step 2 implicitly uses the forward-backward algorithm for the computation of $\gamma^t(i)$, which is traditionally used to compute log-likelihoods  (specifically, the forward part). 

\subsection{Model initialization}

As seen in the previous section, the more complex the context-specific BNs are, the greater the computational complexity of  the learning and inference algorithms.
At the start of training it is therefore assumed that  all the BNs are naïve BNs, i.e., $\kappa_{im}=0$ and $p_{im}=0$ for all variables and hidden states. 
The parameters $\{h_{im}\}_{i=1,m=1}^{N,M}$ are set following the rule of thumb provided by \cite{1986:silverman}: $h_{im} =(4\hat{\sigma}^5_m/(3T))^{\frac{1}{5}}$, $i=1,...,N$, where $\hat{\sigma}_m$ is the sample standard deviation of variable $X_m$.
Each parameter $\{\omega_{il}\}_{i=1,l=P^*}^{N,T}$ is initialized using random draws from a uniform distribution on $[0.4, 0.6]$, followed by normalization to satisfy the constraint  $\sum_{l=P^*}^T\omega_{il} = 1$.  
The $\boldsymbol{\pi}$ parameter is initially assumed to be a uniform categorical distribution, whereas the parameter $\textbf{A}$ is set as a matrix whose diagonal is  filled with value arbitrary far from $1$ (in our case $999$), and the remaining values are $1/N$. 
The matrix is then normalized to be an actual transition matrix. 
This matrix can clearly be modified as needed in case of left-to-right transition matrices or uniform ones. 
The initialization described here is used to condition the model to look for stable patterns. 

\section{Experiments}\label{sec:experiments}

For the experiments, synthetic data and real stochastic data from ambient audio and CNC machines was used to demonstrate the abilities of the proposed model.
Our  model was compared against a traditional HMM, where all the variables are assumed to be independent Gaussians. 
Since the proposed model can be seen as a KDE extension of the AR-AsLG-HMM model in \cite{2022:puerto-santana.larrnaga.ea}, that model was also included in the  comparisons.
In \cite{2022:puerto-santana.larrnaga.ea}, AR-AsLG-HMM showed similar or better results when compared with MoG-HMMs, even when synthetic MoG data was considered. 
Therefore, MoG-HMMs was not included in the experiments. 

If structural optimization is not performed, our KDE-AsHMM model can be seen as a multivariate version of the \cite{2007:piccardi.perez} model when all variables are independent, therefore this model, in the case of assuming independent multivariate data, was also compared.
We denote this model KDE-HMM.
Finally, ablations of the proposed model were also considered for the synthetic data: KDE-BNHMM is the model where BNs were built with no AR structural optimization, while KDE-ARHMM is the model where only AR structural optimization was performed.
Also, a model called KDE-AsHMM* was compared.
This model was provided with the ground truth about AR and non-AR dependency structure, and thus represents performance that might not be attainable in practice when the structure is unknown. 
The differences between the models are stated in Table~\ref{table:model_diff}.

\begin{table}[h]
	\centering
	\begin{tabular}{lcccc}
		Model        & Kernel-based & AR & Non-AR & Ground-truth \\
		\hline
		HMM         & -          & -          & -           & -          \\
		AR-AsLG-HMM & -          & \checkmark & \checkmark  & -          \\
		KDE-HMM     & \checkmark & -          & -           &-           \\
		KDE-AsHMM   & \checkmark & \checkmark & \checkmark  &-           \\
		KDE-ARHMM   & \checkmark & \checkmark & -           &-           \\
		KDE-BNHMM   & \checkmark & -          & \checkmark  &-           \\
		KDE-AsHMM*  & \checkmark & \checkmark & \checkmark  & \checkmark \\
	\end{tabular}     
	\caption{Differences in the models used for validation}
	\label{table:model_diff}
\end{table}

\subsection{Synthetic data}

\subsubsection{Data description}

\begin{figure}[h!]
	\centering
	\begin{tabular}{cc}
		\includegraphics[scale=0.42]{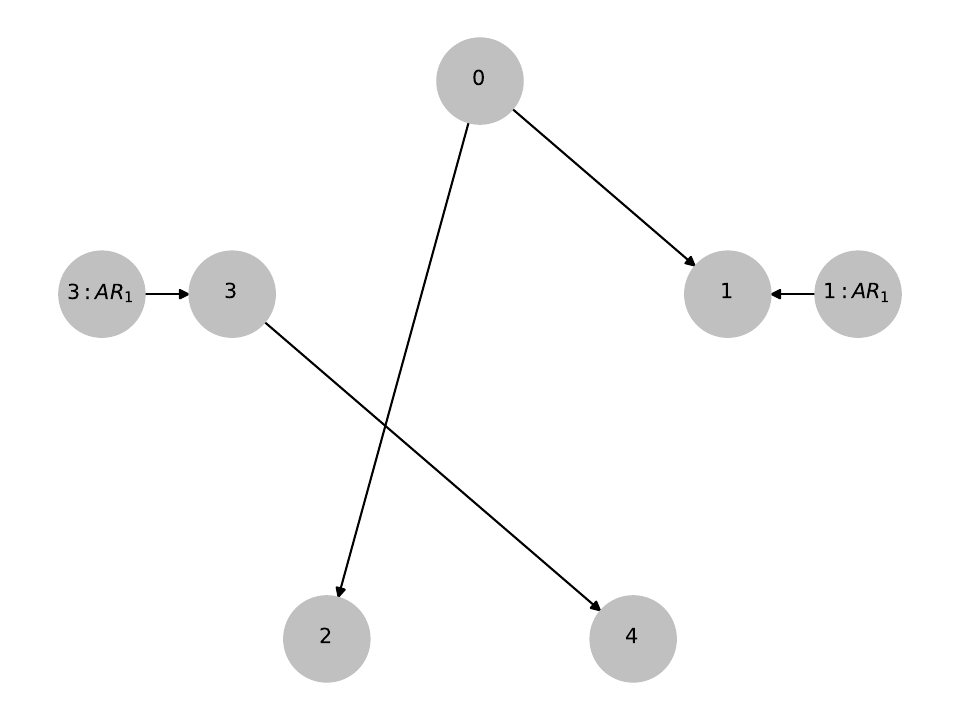} &
		\includegraphics[scale=0.42]{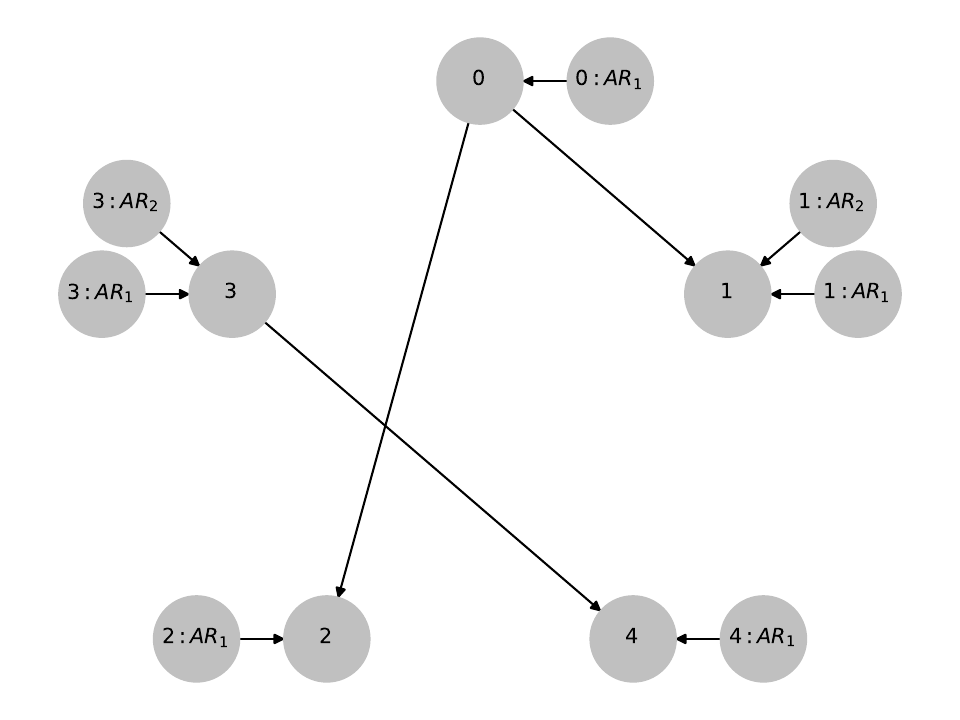} \\
		(a) & (b) \\
	\end{tabular}
	\caption{The context specific graphs corresponding to the synthetic data.
		(a) is a simpler version of the Bayesian network pictured in (b)}
	\label{fig:BNs}
\end{figure}

For the synthetic data, seven variables were used. 
It was assumed that the data jumps between three hidden states, and each hidden state having its own context-specific BN representation as pictured in Figure~\ref{fig:BNs}.
One of the states was assumed to be associated with a naïve Bayes model, hence its graph is not pictured.
The names of the variables go from $0$ to $6$; however, $X_5$ and $X_6$ were assumed to be Gaussian noise which turned them independent of the hidden state variable $Q^t$, and therefore, they were not related to any variable for any hidden state. 
We use  $m : \mathrm{AR}_r$ to denote an arbitrary variable in the network:, is the $r$ AR order of the variable $X^t_m$ or $X^{t-r}_m$. 

\begin{figure}[h]
	\centering
	\includegraphics[scale=0.325]{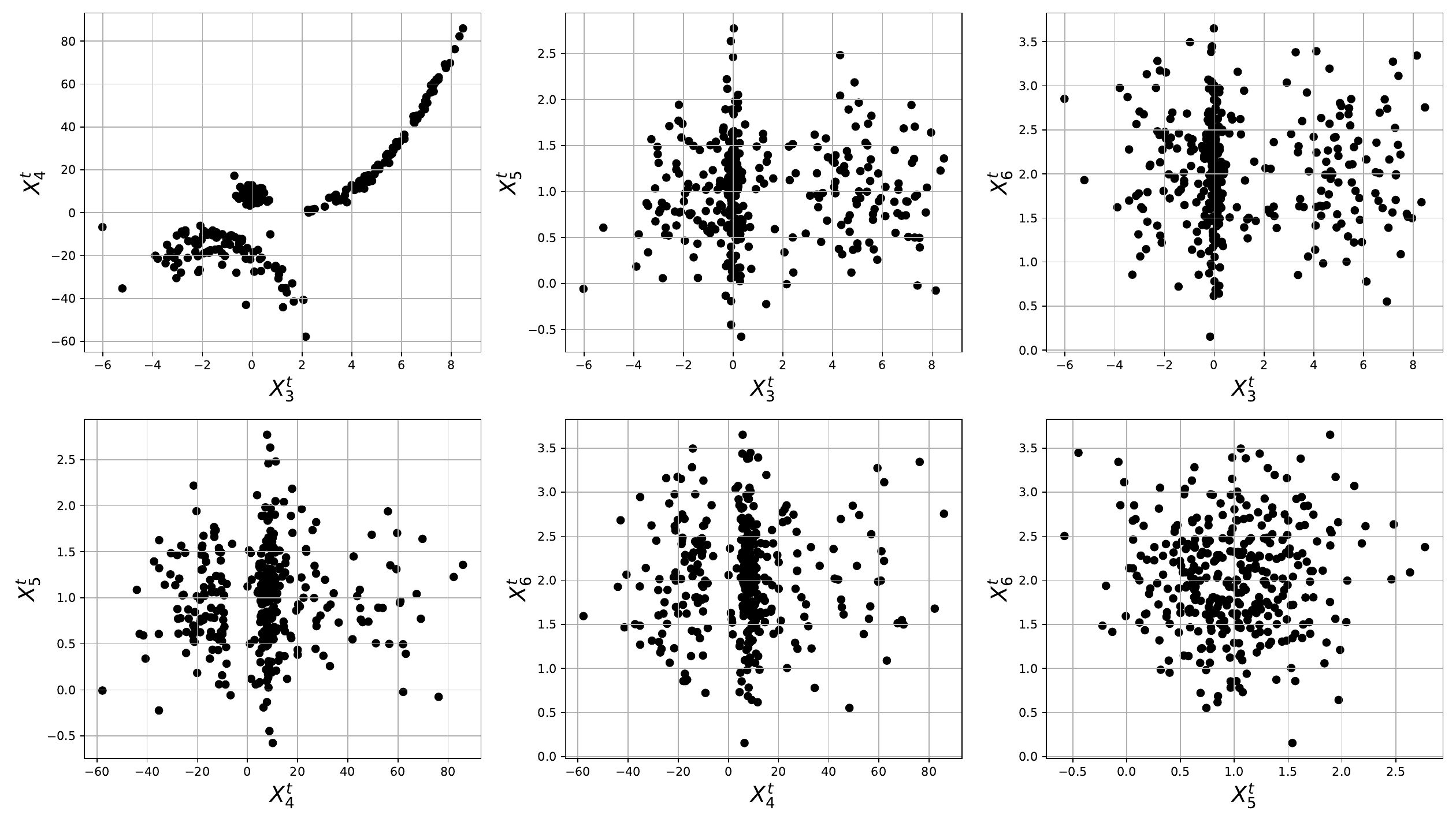}
	\caption{Scatter plots of pairs of variables from synthetic data}
	\label{fig:scatter_data}
\end{figure}

It was assumed that each variable $X_m$ had the following non-linear Gaussian distribution:
\begin{equation}\label{eq:synt_dist}
	X^t_m|Q^t=i \sim \mathcal{N}\left(\sum_{k=1}^{\kappa_{im}}c_{imk}((V^t_{imk})^2-e_{im})+\sum_{r=1}^{p_{im}}d_{imr}X^{t-r}_m,\sigma_{im}^2\right)
\end{equation}
The coefficients $\{ \{c_{imk}\}_{k=1}^{\kappa_{im}},\{d_{imr}\}_{r=1}^{p_{im}},e_{im},\sigma_{im}\}_{i=1,m=1}^{N,M}$ are provided in the complementary material. 
Although the synthetic data was generated using a Gaussian distribution, the non-linear dependencies lead to data behavior that is not modelled well by traditional models.
In Figure~\ref{fig:scatter_data}, scatter plots of pairs of variables from synthetic generated data are pictured.
Being specific, variables $X^t_3$, $X^t_4$, $X^t_5$ and $X^t_6$ are used of the scatter plots.
Note that for some pair of variables, Gaussian behavior is observed, such as the pair $X_5^t-X_6^t$ (since both variables are stationary Gaussian noise), where an ellipsoidal data cluster is observed.
However, for some other pairs, Gaussian distributions with non-linear dependencies are observed. 
As instance, for the pair $X_3^t-X_4^t$, data lay on two quadratic curves.
Each curve in the data description is a hidden state. 
Note that in the scatter plot of the pair $X_3^t-X_4^t$, there is also a small ellipsoidal cluster close to the origin.
The corresponding  data to this cluster matches with the hidden state where a Naive BN is used.
Regarding the remaining plots, observe that the scatter plots are done between Gaussian data with non-linear and linear dependencies.
As instance, in the case of the pair $X_4^t-X_5^t$, Gaussian elliptical clusters are formed around different ranges of the $X_4^t$ axis. 
This implies that the non-linear component of $X^t_4$ is independent of $X^t_5$.

\begin{figure}[h]
	\centering
	\includegraphics[scale=0.45]{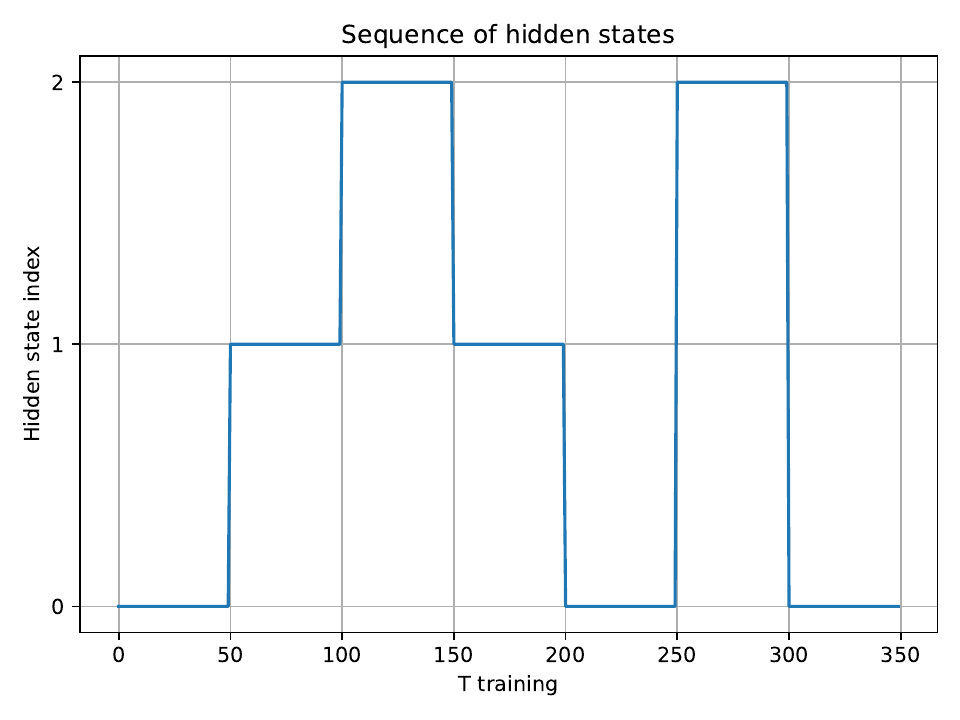}
	\caption{Sequence of hidden states used to generate training and testing data.
		The length of this sequence is expanded proportionally as needed for the training data, maintaining the pattern}
	\label{fig:sequence_hmm}
\end{figure}

The data was generated from a pre-defined sequence of hidden states which is pictured in Figure~\ref{fig:sequence_hmm}.
For the training process, in order to observe how $T$, the amount of training data, affects model performance, the previous sequences was scaled as needed to obtain datasets of length $T\in\{350,700,1050,1400,1750,2100,2450\}$.  
For each $T$, a single time series of that length was sampled and used as training data.
Later, one hundred samples with $T_{\text{test}}=1400$  were generated and used as test dataset. 
To compare the models, the mean log-likelihood  per unit datum and its standard deviation on the testing data were reported.

\subsubsection{Results}

Table~\ref{table:synthetic_results1}  reports the mean and standard deviation of the log-likelihood per unit datum on the testing data.
Note that, when the length of the training data was small ($T=350$), linear models as the AR-AsLG-HMM and HMM could outperform the log-likelihoods reached by the corresponding KDE models such as KDE-AsHMM and KDE-HMM. 
Nonetheless, when the length of the training data sequence increased to $T=2450$, it was observed that all the kernel models obtained a better mean log-likelihood per unit data than the linear models, with the exception of the KDE-HMM. 
This shows that the introduction of context-specific BNs into the kernel models is beneficial to improve how their fitness scales to longer sequences.     
Regarding the standard deviation of such log-likelihoods per unit data, it was observed that AR-AsLG-HMM and HMM obtained the lowest values, indicating that their fitness values were more stable than those obtained by kernel-based HMMs.

\begin{table}[h]
	\centering
	\begin{tabular}{lrrrrrr}
		& \multicolumn{3}{c}{$T=350$} &  \multicolumn{3}{c}{$T=2450$} \\
		\cmidrule(lr){2-4}\cmidrule(lr){5-7}
		Model        &  $\mu$ & $\sigma$ & s & $\mu$ & $\sigma$ &s\\
		\hline
		HMM         &-10.52 & 0.16 & \textbf{0.24}&-10.36 & 0.10&   \textbf{2.07}\\
		AR-AsLG-HMM & \textbf{-9.05} & \textbf{0.11} & 3.64&-8.82  & \textbf{0.09}&   9.98\\
		KDE-HMM     &-12.07 & 0.70 & 1.81& -9.79 & 0.11& 367.15\\
		KDE-AsHMM   & -9.68 & 0.24 &33.60&\textbf{-8.11}  & 0.24&3351.38\\
		KDE-ARHMM   & -9.74 & 0.40 &17.69&-8.17  & 0.25&2229.66\\
		KDE-BNHMM   & -9.95 & 0.47 &62.53&-8.17  & 0.26&5160.31\\
		KDE-AsHMM*  & -9.52 & 0.39 & 7.46&-8.13  & 0.31& 827.93\\
	\end{tabular}     
	\caption{Mean log-likelihood per unit datum  and its standard deviation on the testing data.
		$T$ refers to the size of the length of the training data.
		Onlyresults for the shortest and longest training sequences.
		``s'' is for seconds.
	}
	\label{table:synthetic_results1}
\end{table}

With respect to the training times, it can be seen that  the Gaussian parametric models were faster than their kernel counterparts.
However, the simplest KDE-HMM even in the large datasets, was not excessively high-time demanding. 
Their asymmetric counterparts, KDE-AsHMMs, KDE-BNHMMs and KDE-ARHMMs required up to 60 times more computation time.
In the case of KDE-AsHMM*, as the network is already given, the structural optimization was omitted and the training times were up to 4 times greater than the KDE-HMM.
Observe that KDE-AsHMM* obtained results close to those obtained by KDE-AsHMM with less computational effort.
In this sense, using expert knowledge to build possible dependency graphs is recommended, such that competitive likelihoods can be attained with less computational cost.

\begin{figure}[h]
	\centering
	\begin{tabular}{cc}
		\includegraphics[scale =0.525]{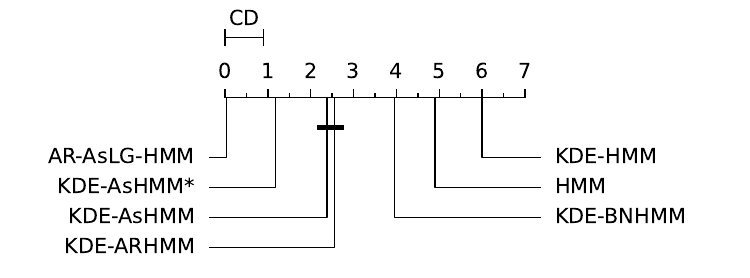}&
		\includegraphics[scale =0.525]{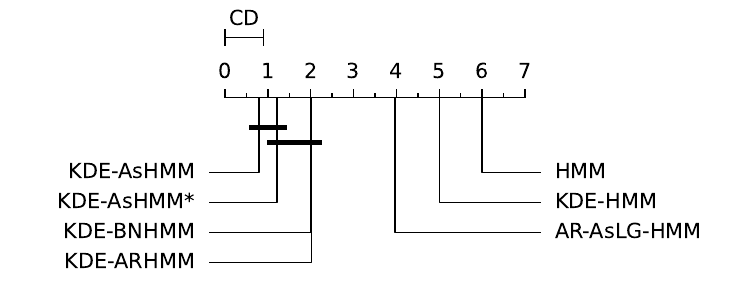} \\
		(a) Nemenyi test $T=350$  & (b) Nemenyi test $T=2450$\\
	\end{tabular}
	\caption{Nemenyi ranking test when (a) the training data length is $T=350$, and (b) the training data length is $T=2450$. Rankings closer to 0 imply a better fit to testing data}
	\label{fig:nemenyi_syn}
\end{figure}

To take into account the effects of variance in the likelihoods and to provide a statistical analysis of the model ranking, the Friedman and the post-hoc Nemenyi hypothesis test was used, see \cite{2006:demsar}.
The Friedman test checks the hypothesis of global no difference in rankings.
If the hypothesis is rejected the post-hoc Nemenyi test is applied. 
For all pairs of models, the method subsequently tests the hypothesis of no difference in mean ranking in terms of the mean log-likelihood per datum.
A critical difference (CD) value  is computed to indicate the minimum distance in the rank needed to declare evidence of statistical difference.

The Friedman test when $T=350$ provided a test statistics of $712.53$ which corresponds to a virtual p-value of $0.0$.
Therefore, the null hypothesis of no difference in ranking between the methods was rejected.
When $T=2450$, the test statistics was $663.63$ which again gave a $p$-value of $0.0$.
Since null hypothesis was rejected in both cases, we proceeded with the Nemenyi post-hoc tests.

The results of the Nemenyi tests are provided in Figure~\ref{fig:nemenyi_syn}.
In (a), the test was applied when the training data length was $T=350$, and it was observed that the top ranked models were AR-AsLG-HMM and KDE-AsHMM*, and the worst were KDE-HMM and HMM.
Note that there is no statistical evidence to claim that KDE-AsHMM and KDE-ARHMM were different in the ranking position.
Nevertheless, in (b), when $T=2450$, the best ranked models were KDE-AsHMM and KDE-AsHMM*, and the worst were again HMM and KDE-HMM. 
However, there was no statistical evidence to claim that KDE-AsHMM and KDE-AsHMM* differed in their ranking. 
On the other hand, the HMM and KDE-HMM models presented clear evidence of differences in their ranking position.

\begin{figure}[h]
	\centering
	\begin{tabular}{cc}
		\includegraphics[scale=0.4]{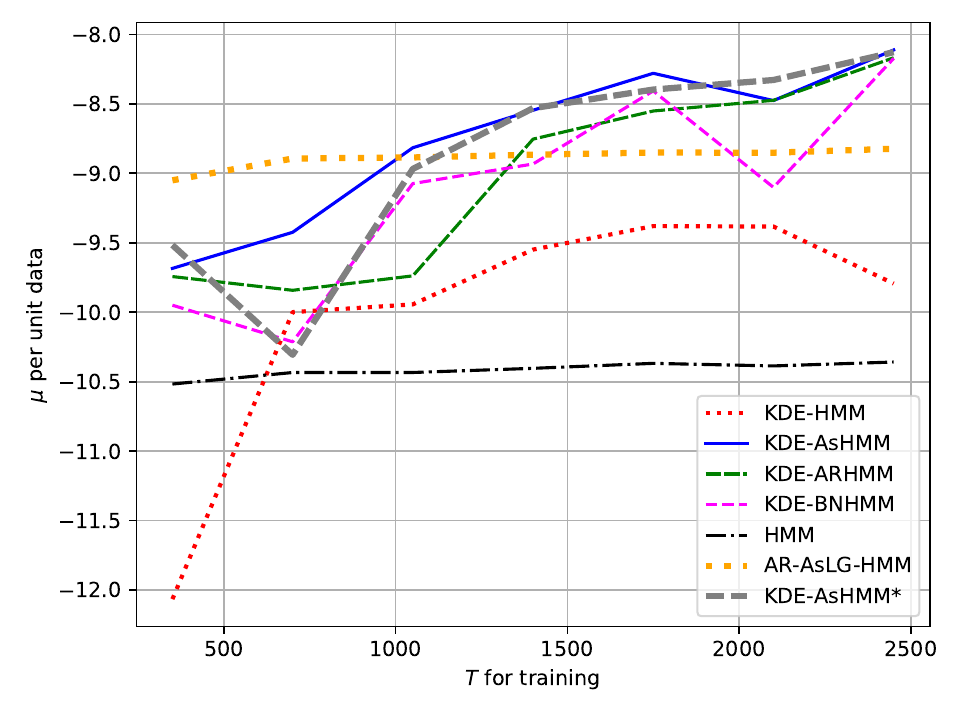}&
		\includegraphics[scale=0.4]{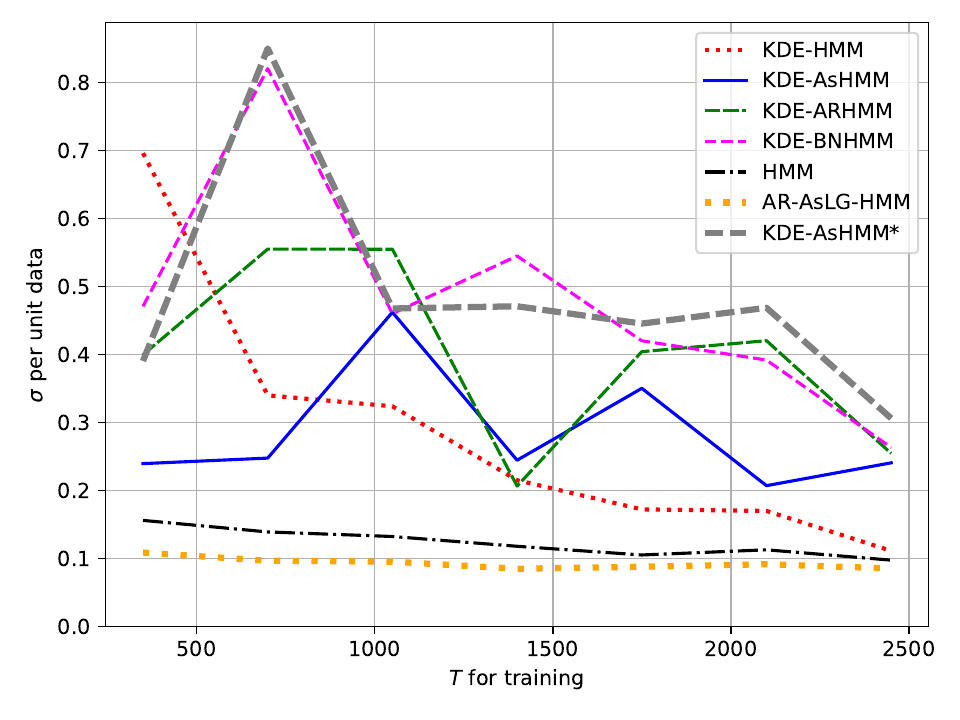}\\
		(a) & (b) \\
	\end{tabular}
	\caption{ In (a) the mean log-likelihoods per datum when the size $T$ of training data changes and in (b) the corresponding standard deviation}
	\label{fig:xunit_ll}
\end{figure}

Figure~\ref{fig:xunit_ll} expands on the previous results in mean and standard deviation of log-likelihood per unit datum.
In this case, the results when the training data length $T\in\{350,700,1050,1400,1750,2100,2450\}$ are reported in plot (a) and (b), where (a) shows the evolution of the mean log-likelihood per unit datum and (b) pictures its standard deviation.
In (a), it was observed that AR-AsLG-HMM was capable of obtaining the best results in terms of likelihood when the training data was small.
However, it is also evident from the plot how the performance of the non-parametric kernel models kept improving as more training data became available, unlike the parametric models (HMM and AR-AsLG-HMM). 
It is worthy to note that, regarding kernel models, which of KDE-AsHMM* and KDE-AsHMM that obtained the best log-likelihood differed for different $T$ values.
Regarding (b), Gaussian parametric models, in spite of not getting benefits in likelihood with increases in training data, they did have lower standard deviation, but the variance reductions were not substantial. 
On the other hand, KDE-HMM greatly reduced variance when $T$ increased, but this did not translate into relevant improvements in terms of log-likelihood, at least, enough to outperform AR-AsLG-HMM.
The proposed model KDE-AsHMM and its ablations, showed slightly reduced variance with increasing $T$, but we do not have statistical evidence to claim that this trend is statistically significant.

To summarize the results on synthetic data, the kernel models in the HMM framework (with the exception of KDE-HMM) all were able to take advantage of the increase in the training dataset size, and the addition of context-specific BNs improved model performance, but longer training times and greater variance in fitness could be observed as well.

\subsection{Real data from environmental sound classification  }

\subsubsection{Data description}
For these experiments,  the \textit{Environmental Sound Classification 50}\footnote{https://www.kaggle.com/datasets/mmoreaux/environmental-sound-classification-50} dataset was used.
The dataset consists of environmental sounds from fifty different sources as cats, dogs, keyboards, snoring, mouse, clicks, etc.
Each sound is recorded at a sample rate of 16~kHz during 5 seconds.  
From the raw audio files, 5 mel frequency spectrum coefficients (MFCCs) \cite{1980:davis.mermelstein} were extracted.
The time window had a range of 0.1 seconds or 1600 time instances and the and the hop length was 0.05 seconds or 800 time instances. 
The dataset was divided into 5 folds, where each fold had eight recordings for each class.
Therefore, for the model validation, a 5-fold cross-validation was performed.
In this use case, we are concerned with identifying which model type, among those considered in these experiments, that obtains the best results in terms of classification accuracy.
Therefore, for each model type and fold, a separate model was trained on each class.
The prediction in testing phase was done by selecting the class of  the model which maximized the log-likelihood of the data.
Since the classes are equally probable a-priori, this is the same as predicting the most probable class according to the class-conditional models, which is theoretically optimal for classification.
In the case of KDE-AsHMM, to prevent long computational times, the SEM algorithm was iterated only once and $P^*$ was fixed as $P^*=1$. 

\subsubsection{Results}

\begin{table}[h]
	\centering
	\begin{tabular}{lrrrrrr}
		Model        &  F1(\%)  & F2(\%)  & F3(\%)  & F4(\%)  & F5(\%)   &mean(\%)  \\
		\hline
		HMM          &22.3 & 18.3  & 18.8 & 26.3 & 18.5 & 20.8  \\
		AR-AsLG-HMM  &28.0 & 29.3  & 29.8 & 30.3 & 31.8 & 29.8\\
		KDE-HMM      &13.5 & 15.8  & 17.0 & 18.5 & 13.3 & 15.6 \\
		KDE-AsHMM    &\textbf{32.3} & \textbf{35.3}& \textbf{36.3 }&\textbf{ 41.8} &\textbf{ 32.3} & \textbf{35.6} \\
	\end{tabular}
	\caption{Fold and mean accuracy for the 50 ambient sounds classification problem }
	\label{table:synthetic_audio}
\end{table}

In this case only four models were compared due to the computational cost: they were HMM, AR-AsLG-HMM, KDE-HMM and KDE-AsHMM.  
The ablations were omitted, since they were found to have equal or worse performance than KDE-AsHMM in the previous experiment. 
Since models must be trained for each class in each fold, and there are 50 classes to be classified, 1000 models had to be learned in total (250 models by model type).
All models we trained used three hidden states to describe the MFCC training data.
A random classification of the testing files would obtain in mean an accuracy of $2\%$.
Therefore, this can be seen as the lowest tolerable or bottom line accuracy for a classifier. 
The results of classification from each model for each fold are provided in Table~\ref{table:synthetic_audio}.
As it can be seen, for all the folds, the model with the highest accuracy was the proposed model KDE-AsHMM, followed by AR-AsLG-HMM, HMM and KDE-HMM.
The latter had  the lowest accuracy, which implies that the addition of BNs for kernel correction and information sharing can be via BNs can be helpful to have more accuracy in this classification problem.
Finally, although the accuracies did not exceed $50\%$, they were at least 5 times higher than the baseline accuracy and at most 20 times higher, which indicates that the models perform substantially better than random prediction.

\begin{figure}[h]
	\centering
	\begin{tabular}{cc}
		\includegraphics[scale=0.4]{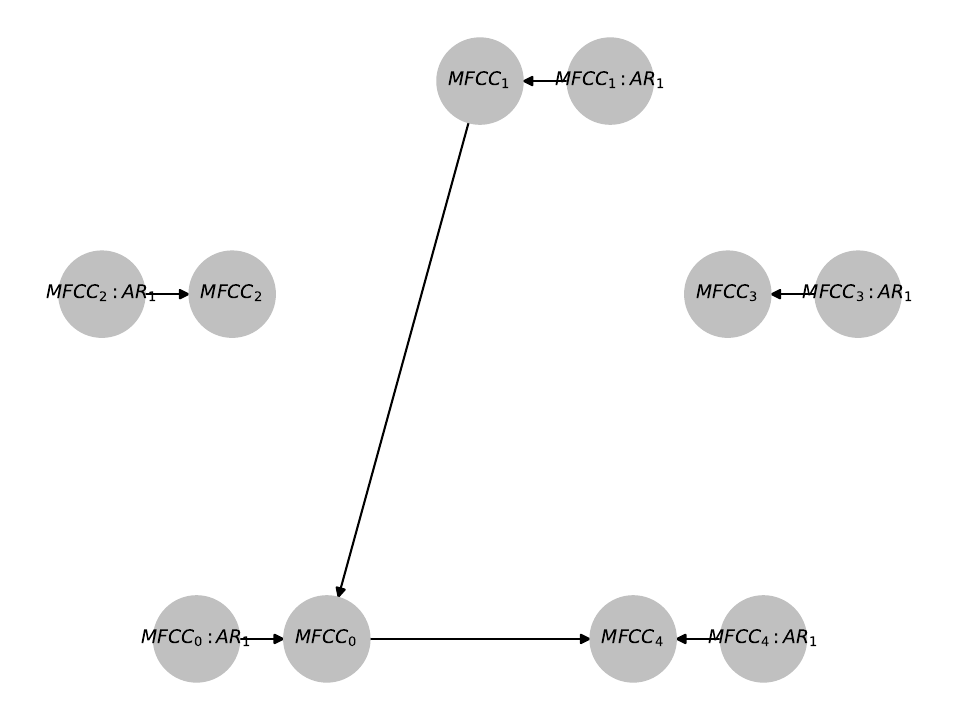}&
		\includegraphics[scale=0.4]{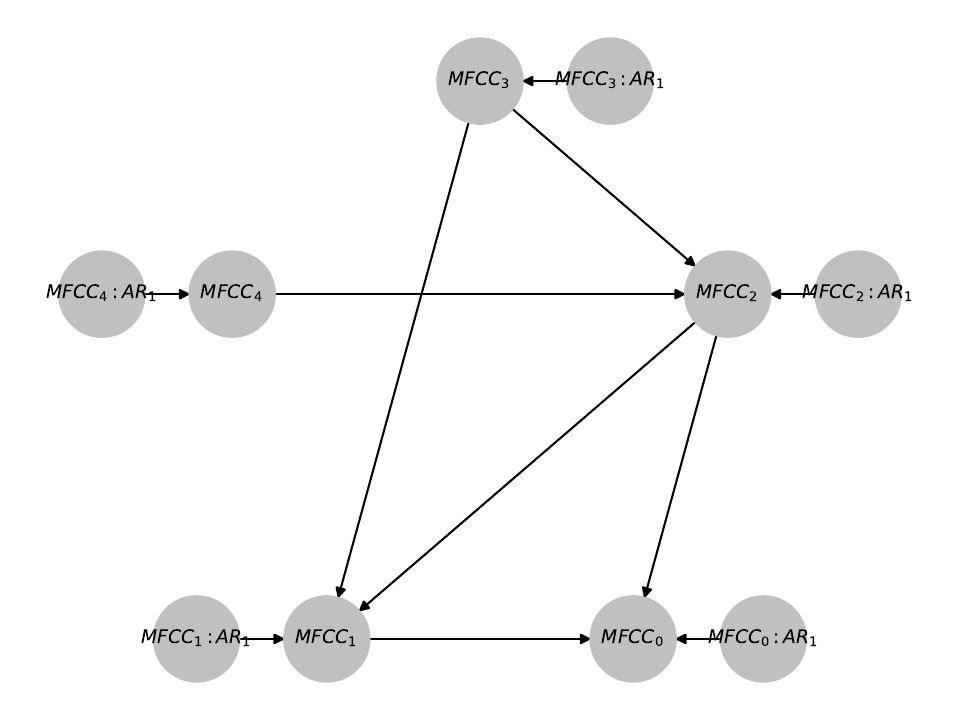}\\
		(a) & (b) \\
	\end{tabular}
	\caption{BNs obtained from the pig class at fold 1 from KDE-AsHMM. (a) and (b) represent two different hidden states}
	\label{fig:BN_pig_audio}
\end{figure}

Recall that the proposed model introduces context-specific BNs into KDE-HMMs, which are used to provide kernel corrections, based on the information from AR values and between-variable dependencies.
In Figure~\ref{fig:BN_pig_audio}, two context-specific BNs from the pig audio KDE-AsHMM are pictured. 
In (a), AR values were present to explain the MFCC amplitudes, this makes sense given the slowly changing nature of most sounds, and some cross-dependencies such as the 4th MFCC depending on the 0th MFCC. 
In  (b), it can be seen that most of the previous relationships hold but further relationships appear, for example, the 2nd MFCC depends on the 4th and 3rd MFCC. 
These relationships can provide further insights from the learned audio and the sound generation for the class instance, in this case, a pig.

\subsection{Real data from a CNC mill tool}

\subsubsection{Data description}
In this section we consider \textit{CNC mill tool wear} dataset\footnote{https://www.kaggle.com/datasets/shasun/tool-wear-detection-in-cnc-mill}.
A series of machining experiments were run on $5.08~\text{cm} \times 5.08~\text{cm} \times 3.81~\text{cm}$ wax blocks in a CNC milling machine in the System-level Manufacturing and Automation Research Testbed (SMART) at the University of Michigan.
Machining data was collected from a CNC machine for a variety  of tool condition, feed rate, and clamping pressure.
Each experiment produced a finished wax part with an ``S" shape carved into the top face.
The data contains samples where the extracted S shaped part achieved the desired quality and others where that level of quality was not reached.
The quality of a part was determined by visual inspections.

\begin{figure}[h]
	\centering
	\includegraphics[scale=0.28]{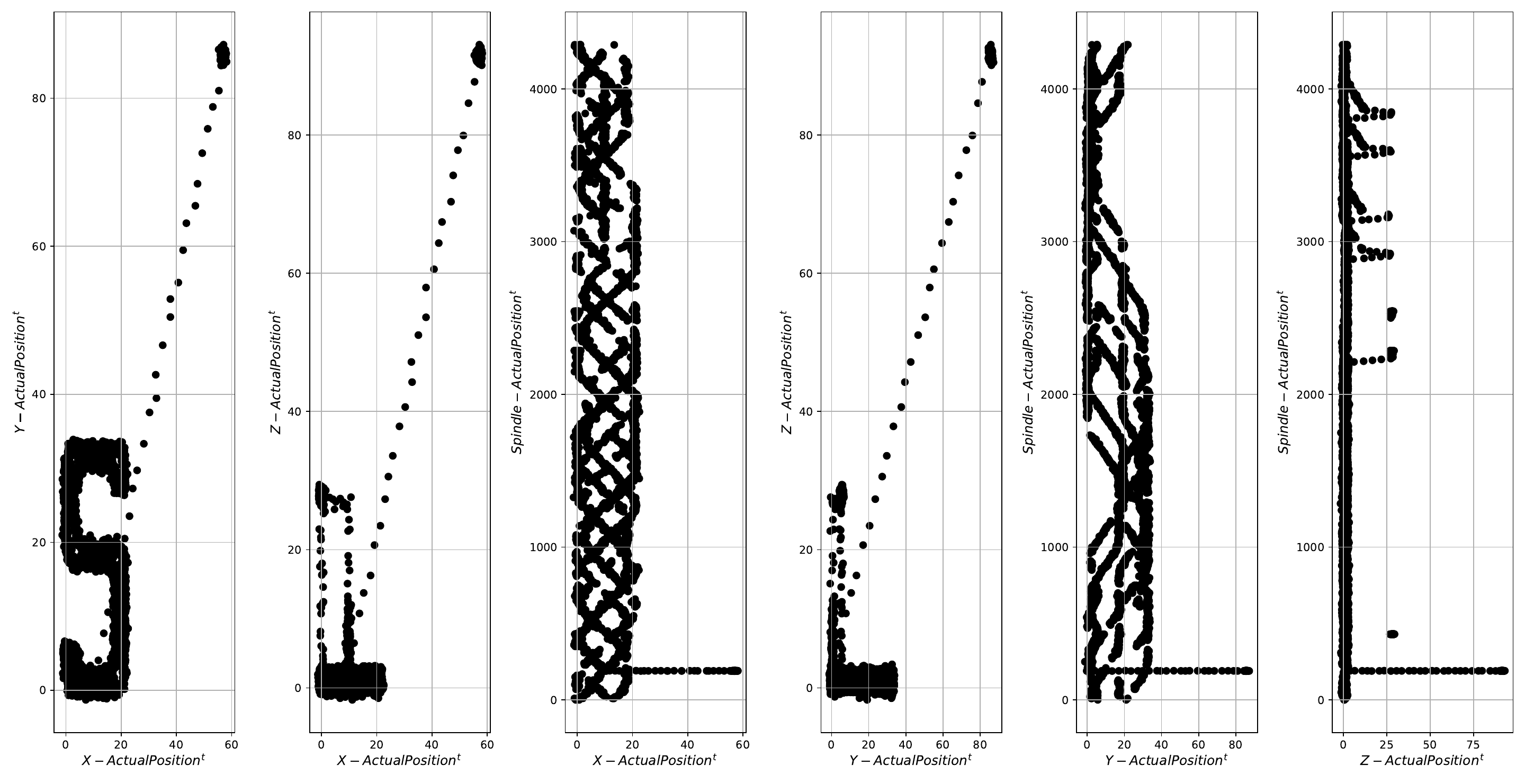}
	\caption{Scatter plots for all pairs of selected features.
		In the X-Y actual position plane, the S-shape extracted piece from the wax block is seen. 
		The remaining plots also show nonlinear relationships, for instance involving X-spindle actual position plane and Y-spindle plane}
	\label{fig:s-shape}
\end{figure}

Table~\ref{table:cnc_des} provides a brief description of the 18 experiments.
Since it is observed that a worn tool can provide an accepted S-piece, we focus our study on the utility of log-likelihoods from different model types for discriminating between accepted and non-accepted pieces.   
The dataset contains 44 features related to  each axis of the part: its position, velocity, acceleration, current, voltage and power; whereas from the spindle, its position, velocity, acceleration, current, voltage, power and inertia were also recorded. 
However, we saw from scatter plots that the variables that exhibited the most complex data distribution came from the features related to the position of the spindle and the piece.
The remaining variables were either constant or closer to Gaussian behavior. 
Therefore, only 4 variables were considered in our experiments, namely: \textit{actual X-axis position of the part}, \textit{actual Y-axis position of the part}, \textit{actual Z-axis position of the part} and \textit{actual position of the spindle}. 
Figure~\ref{fig:s-shape}  provides scatter plots for all pairs of variables we included. 
Note how, in the X-Y actual position plane, the S-shape figure is observed.

\begin{table}[h]
	\begin{tabular}{lllll}
		\textbf{Essay} & \textbf{Tool condition} & \textbf{Experiment ended?} & \textbf{Accepted?} & \textbf{Length}\\
		\hline
		1& unworn &yes&yes & 1055\\
		2& unworn &yes &yes & 1668\\
		3& unworn &yes &yes & 1521\\
		4& unworn &no   &no  & 532 \\
		5& unworn &no   &no  & 462 \\
		6& worn   &yes  &no  & 1296\\ 
		7& worn   &no   &no  & 565 \\
		8& worn   &yes  &no  & 605 \\
		9& worn   &yes  &no  & 740 \\
		10&worn   &yes  &no  & 1301\\
		11&unworn &yes  &yes & 2314\\
		12&unworn &yes  &yes & 2276\\
		13&worn   &yes  &yes & 2233\\
		14&worn   &yes  &yes & 2332\\
		15&worn   &yes  &yes & 1381\\
		16&worn   &no   &no  & 602 \\
		17&unworn &yes  &yes & 2150\\
		18&worn  &yes  &yes & 2253\\
	\end{tabular}
	\caption{Description for \textit{CNC mill tool wear} dataset}
	\label{table:cnc_des}
\end{table}
Observe that the experiments 1, 2, 3, 11, 12, 13, 14, 15, 17 and 18 yielded an accepted piece.
From these 10 sequences, 5 folds were created. 
Namely, \textit{Fold 1} used essays 1 and 13, \textit{Fold 2} used 2 and 14, \textit{Fold 3} used 3 and 15, \textit{Fold 4} used 11 and 17, \textit{Fold 5} uses 12 and 18. 
We trained one model each of HMM, AR-AsLG-HMM, KDE-HMM and KDE-AsHMM on the sequences in each fold.
The instances of the other four folds were used for testing.
This allowed us to assess how well these models fit sequences that yielded accepted pieces.
On the other hand, the essays 4, 5, 6, 7, 8,  9,  10 and  16 produced non-accepted pieces.
All these instances were evaluated for all the folds and models, so that we can compare the difference in model fitness between sequences that yielded accepted and non-accepted pieces.
For fitness, the log-likelihood per unit datum was used and reported.
Like in our previous experiment, we set $P* = 1$ and only iterated the SEM structure search once

In each dataset, every time instance is also labeled with a processing state.  
Among those, we used the following labels for building and initializing models: `Layer 1 Down', `Layer 1 Up', `Layer 2 Up', `Layer 2 Down', `Layer 3 Down' and `Layer 3 Up'.
As there are 6 processing states, 6 hidden states were used in each model.
The process-state annotations were also used to determine the initial values of the $\boldsymbol{\omega}$ parameters.
Assume that a matching between hidden states and machining states  is created. 
Then, set $\omega_{il}=1$ if  the machining state of the instance $l$ correspond to the model state $i$, otherwise, set $\omega_{il} = 1e-5$. 
Next, normalize $\boldsymbol{\omega}$ in order to obtain a valid initial parameter value.

\subsubsection{Results}

\begin{table}
	\begin{tabular}{lrrrrrr}
		& \multicolumn{3}{c}{\textbf{Fold 1}} & \multicolumn{3}{c}{\textbf{Fold 2}} \\
		\cmidrule(lr){2-4}\cmidrule(lr){5-7}
		Model      & \textit{Good}    & \textit{Bad}      &\textit{Diff.}   & \textit{Good}    & \textit{Bad}     & \textit{Diff.}\\
		HMM         & -253.08 & -507.18 &\textbf{-254.10}&-20.39&-103.49&-83.10\\ 
		AR-AsLG-HMM & -394.28 & -334.16 &60.12&-35.17&-283.31&\textbf{-248.15}\\ 
		KDE-HMM     & -334.79 & -341.14 &-6.35&-176.60&-353.75&-177.14\\ 
		KDE-AsHMM   & \textbf{-9.49} & -22.31 &-12.82&\textbf{-18.34}&-61.85&-43.51\\ 
		& \multicolumn{3}{c}{\textbf{Fold 3}} & \multicolumn{3}{c}{\textbf{Fold 4}} \\
		\cmidrule(lr){2-4}\cmidrule(lr){5-7}
		Model      & \textit{Good}    & \textit{Bad}      &\textit{Diff.}   & \textit{Good}    & \textit{Bad}     & \textit{Diff.}\\
		HMM         & -16.22 & -22.07 &-5.85&-30.10&-46.52&-16.42\\ 
		AR-AsLG-HMM & \textbf{-6.98} & -49.15 &-42.17&\textbf{-16.09}&-29.70&-13.61\\ 
		KDE-HMM     & -69.27 & -119.83 &\textbf{-50.56}&-271.05&-351.14&\textbf{-80.08}\\ 
		KDE-AsHMM   & -10.31 & -33.65 &-23.35&-47.01&-37.03&9.98\\ 
		& \multicolumn{3}{c}{\textbf{Fold 5}} & \multicolumn{3}{c}{\textbf{Mean}} \\
		\cmidrule(lr){2-4}\cmidrule(lr){5-7}
		Model      & \textit{Good}    & \textit{Bad}      &\textit{Diff.}   & \textit{Good}    & \textit{Bad}     & \textit{Diff.}\\
		HMM         & -23.76 & -46.70 &-22.94&-68.71&-145.19&\textbf{-76.48}\\ 
		AR-AsLG-HMM & -17.10 & -73.74 &\textbf{-56.64}&-93.92&-154.01&-60.09\\ 
		KDE-HMM     & -29.85 & -42.14 &-12.29&-176.31&-241.60&-65.28\\ 
		KDE-AsHMM   & \textbf{-5.08} & -10.58 &-5.50&\textbf{-18.05}&-33.08&-15.04\\ 
	\end{tabular}
	\caption{Log-likelihood per datum for each model and fold.
		The columns corresponding to \textit{Good},(\textit{bad}) refer to the mean fitness of the models regarding accepted (non-accepted) pieces.
		The columns corresponding to \textit{Diff.}\ refer to the difference \textit{Bad}--\textit{Good}.
		For the set of results in \textit{Mean}, the mean values across the folds are averaged}
	\label{table:cnc_results}
\end{table}

Table~\ref{table:cnc_results}  shows the experimental results, in terms of the log-likelihood per datum obtained for each fold.
Three columns of numbers are provided for each fold, namely: \textit{Good} which refers to the mean fitness obtained on the time series from accepted pieces in other folds, \textit{Bad} which refers to the mean fitness obtained when evaluating the same model on the non-accepted-pieces time series, and \textit{Diff.}, which is the result of subtracting \textit{Good} of \textit{Bad}. 
This last column can be interpreted as a measure of the quality of the model to differentiate between accepted pieces and non accepted pieces.
Thus, negative values, implies that the model is able to determine a non-acceptable piece. 
%
%
As comment, recall that there is no model for non-accepted-pieces and is not possible to make per-sequence differences to evaluate the models.
The motivation for this, is that the failure mode from a process can generate unbalanced data sets \cite{2020:diaz-rozo.bielza.ea}, and to obtain enough data to model each failure mode can be expensive to gather or unavailable.

From the results, we see that our proposed approach obtained the best likelihood for every fold when it comes to explaining and modeling processes leading to accepted pieces.
Regarding the difference results, or the ability of the models to differentiate between processes that lead to accepted parts and non-accepted parts, it is observed that for most of the folds, all the models are capable of differentiating between acceptable and non-acceptable pieces for all folds, except in Fold 1 where  AR-AsLG-HMM obtained results above zero, and in Fold 4 where KDE-AsHMM obtained a difference above zero. 
In the mean column, it is observed that in overall, the KDE-AsHMM obtained the best results in terms of fitness followed by HMM and AR-AsLG-HMM.
Regarding the differences between log-likelihoods of accepted and non-accepted pieces, the greatest difference was also obtained by HMM, followed by KDE-HMM and AR-AsLG-HMM.
This implies that our model was the best to model the data from accepted pieces.
It also generalized well to data from non-accepted pieces, achieving the highest log-likelihoods in mean there as well. 
Conversely, that good performance meant that the model did not produce as strong a contrast between accepted and non-accepted pieces as other models, but it still achieved a difference below zero on average.

\begin{figure}[h]
	\centering
	\includegraphics[scale=0.8]{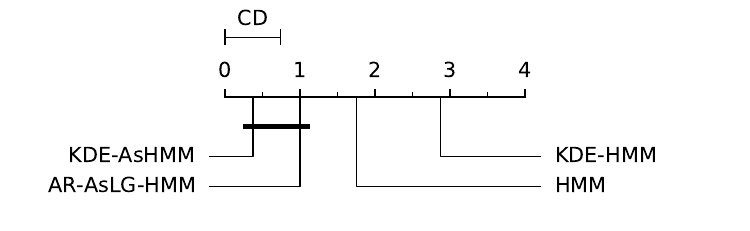}
	\caption{Nemenyi hypothesis testing for ranking positions regarding the log-likelihood obtained on sequences from accepted pieces.
		The closer to zero, the better the fitness obtained by the model.}
	\label{fig:nemenyi_cnc}
\end{figure}

To test for statistically significant differences in ranking positions for the four models the Friedman test was applied.
To perform the test, the four different models were ranked in terms of their log-likelihood on each held-out accepted-piece sequence for every fold. 
This led to a total of $8 \times 5 = 40$ ranked lists of the four different models.
In this case the test statistic value is $179.25$, which leads to a virtual $p$-value of $0.0$. 
This means that the null-hypothesis could be rejected and the Nemenyi post-hoc test was applied.
The results of that  test are pictured in Figure~\ref{fig:nemenyi_cnc}, where it is observed that KDE-AsHMM obtained the best mean ranking, followed by AR-AsLG-HMM and HMM. 
From the test, the models that were statistically different in their rankings were HMM and KDE-HMM.
Nonetheless, there was not enough statistical evidence to determine that KDE-AsHMM obtained a different rank position to AR-AsLG-HMM.

\begin{figure}[h]
	\centering
	\begin{tabular}{cc}
		\includegraphics[scale=0.4]{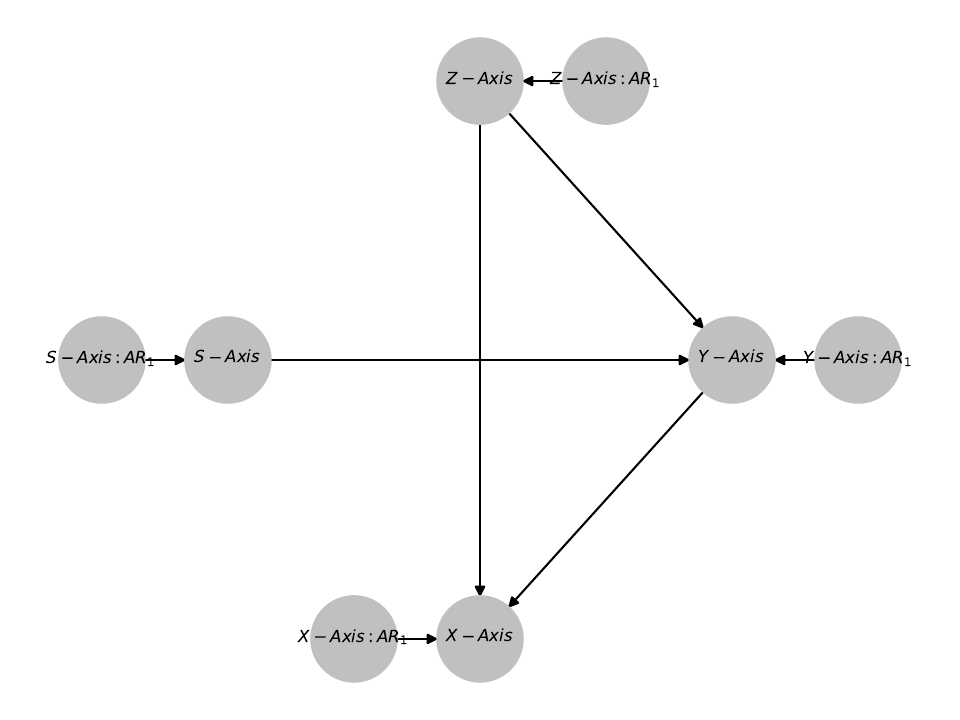}&
		\includegraphics[scale=0.4]{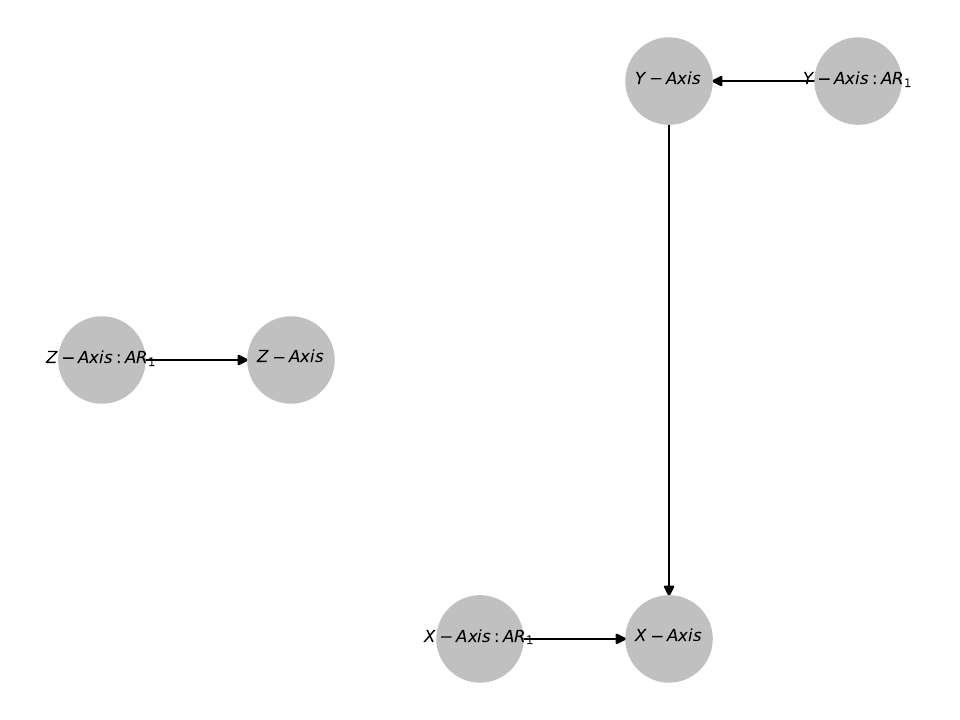}\\
		(a) & (b) \\
	\end{tabular}
	\caption{BNs obtained from the \textit{CNC mill tool wear} data}
	\label{fig:BN_cnc_machine}
\end{figure}

Finally, recall that our proposed model is capable of providing BNs that can be used to understand and generate further insights regarding the data.
In this case, the BNs obtained  for fold 3 are illustrated in Figure~\ref{fig:BN_cnc_machine}.
To save space, we only show diagrams for two of the six model states.
In (a) we find a BN where all the axes rely on their AR values, values and the X, Y and Z axes all are interrelated, whereas the spindle axis is only directly related to the Y axis
Meanwhile in (b), it is observed that in this case the  X, Y and Z axes depend on AR values.
The spindle axis is not statistically related to any other axis position not its AR values.
Nonetheless, the X axis is statistically related to the Y axis. 
This implies that depending on the hidden state some axes may be independent of others and AR values can be relevant to explain the current behavior.
\section{Conclusions}\label{sec:conclusions}

This article introduced a new kind of asymmetric hidden Markov model.
The model introduces kernel corrections via conditional dependencies represented using context-specific Bayesian networks.
We have provided a learning procedure based on the EM  and SEM algorithms.
Additionally, we have provided theoretical computational bounds for the learning and inference models in terms of big $O$ notation, both for dense and naïve Bayesian networks.
From these results, it was noted that, for the computationally heaviest phases of the learning algorithm, the complexity depends on the square of the length of the input data, and on the  fourth-power of the feature space dimensionality.

We have performed experiments on synthetic and real data, showing how the proposed models benefited from using more data in the training phase.
It was observed that parametric models could outperform kernel based ones on our synthetic data; nonetheless, when enough data is available, the proposed model, thanks to its flexibility provided by the kernel-corrections, was capable of excel in terms of log-likelihood per unit datum.
It was also noted that computation time can increase substantially during the search of structures for the model kernel-correction mechanism.
As a consequence, it is advised to provide a limited set of context-specific Bayesian networks for the models to search over.

For the applications to real-life data, we studied model accuracy for classification on a dataset from ambient sound classification. 
The bottom line accuracy was 2\% (50 classes), whereas the proposed model was observed to achieve accuracies up to 45\%, which outperformed other tested HMMs.
It was also observed that for some cases, parametric models could perform better than a kernel based model without kernel-corrections.
This again highlights the utility of enabling  information sharing among the features for a model. (In this case, the information is shared using context-specific Bayesian networks).
Finally, data from a CNC  machine  was studied, where a spindle extracted an S-shape piece from a wax block.
The proposed model obtained the best results in terms of log-likelihood per unit datum and was also able to detect non-acceptable pieces.  
Also, the learned BNs could provide further data insights in the form of statistical dependencies between axes during the machining process.

In other articles such as \cite{2022:atienza.bielza.ea}, it was observed that better modeling accuracy resulted in the Bayesian network structure, when the nodes were allowed to change from linear models to kernel density estimation models. 
As future work, it thus seems relevant to investigate how these transformations can be fitted in our proposed model and how it can be beneficial to reduce the computational time burdens.
Additionally, the authors believe that the current implementation can be improved in order to reduce training time.

\section*{Acknowledgments}
This work was partially supported by the Spanish Ministry of Science and Innovation through the PID2019-109247GB-I00 and DSTREAMS project RTC2019-006871-7. This work was partially supported by the Wallenberg AI, Autonomous Systems and Software Program (WASP) funded by the Knut and Alice Wallenberg Foundation.

\bibliographystyle{ieeetr} 
\bibliography{last_version}

\end{document}